\pdfoutput=1

\documentclass[11pt]{article}

\usepackage[]{acl}

\usepackage{times}
\usepackage{latexsym}

\usepackage[T1]{fontenc}

\usepackage[utf8]{inputenc}

\usepackage{microtype}

\usepackage{inconsolata}

\usepackage{tikz}
\usepackage{bbm,bm}
\usepackage{booktabs}
\usepackage{amsmath,amssymb,amsfonts}
\usepackage[normalem]{ulem}
\useunder{\uline}{\ul}{}
\usepackage{color, colortbl}
\usepackage{xspace}
\usepackage{listings}
\usepackage{microtype}
\usepackage{enumitem}
\usepackage{setspace}
\usepackage{amssymb,amsmath}
\usepackage{makecell}

\usepackage{amsmath}
\usepackage{booktabs}
\usepackage{multirow}

\newcommand{\fairdata}{\textsc{PerspectiveSumm}\xspace}

\title{Fair Abstractive Summarization of Diverse Perspectives}

\author{Yusen Zhang$^{\clubsuit}$ 
\quad Nan Zhang$^{\clubsuit}$ 
\quad Yixin Liu$^\dagger$ \\
\bf \quad Alexander Fabbri$^\diamondsuit$
\bf \quad Junru Liu$^\spadesuit$  
\bf \quad Ryo Kamoi$^\clubsuit$
\bf \quad Xiaoxin Lu$^\clubsuit$ \\
\bf \quad Caiming Xiong$^\diamondsuit$  
\bf \quad Jieyu Zhao$^\ddagger$ 
\bf \quad Dragomir Radev$^\dagger$ 
\bf \quad Kathleen McKeown$^\heartsuit$ 
\bf \quad Rui Zhang$^\clubsuit$ \\
  $^\clubsuit$Penn State University \quad
  $^\dagger$Yale University \quad
  $^\diamondsuit$Salesforce Research\quad  \\
  $^\spadesuit$Texas A\&M University \quad
  $^\ddagger$University of Southern California \quad
  $^\heartsuit$Columbia University \quad
\\
\texttt{\{yfz5488, rmz5227\}@psu.edu}  
  } 

\begin{document}
\maketitle
\begin{abstract}
People from different social and demographic groups express diverse perspectives and conflicting opinions on a broad set of topics such as product reviews, healthcare, law, and politics. A fair summary should provide a comprehensive coverage of diverse perspectives without underrepresenting certain groups.
However, current work in summarization metrics and Large Language Models (LLMs) evaluation has not explored fair abstractive summarization.
In this paper, we systematically investigate fair abstractive summarization for user-generated data. 
We first formally define fairness in abstractive summarization as not underrepresenting perspectives of any groups of people, and we propose four reference-free automatic metrics by measuring the differences between target and source perspectives.
We evaluate nine LLMs, including three GPT models, four LLaMA models, PaLM 2, and Claude, on six datasets collected from social media, online reviews, and recorded transcripts. Experiments show that both the model-generated and the human-written reference summaries suffer from low fairness. We conduct a comprehensive analysis of the common factors influencing fairness and propose three simple but effective methods to alleviate unfair summarization. 
Our dataset and code are available at \url{https://github.com/psunlpgroup/FairSumm}.
\end{abstract}

\begin{figure}[!th]
\centering
\includegraphics[width=\linewidth]{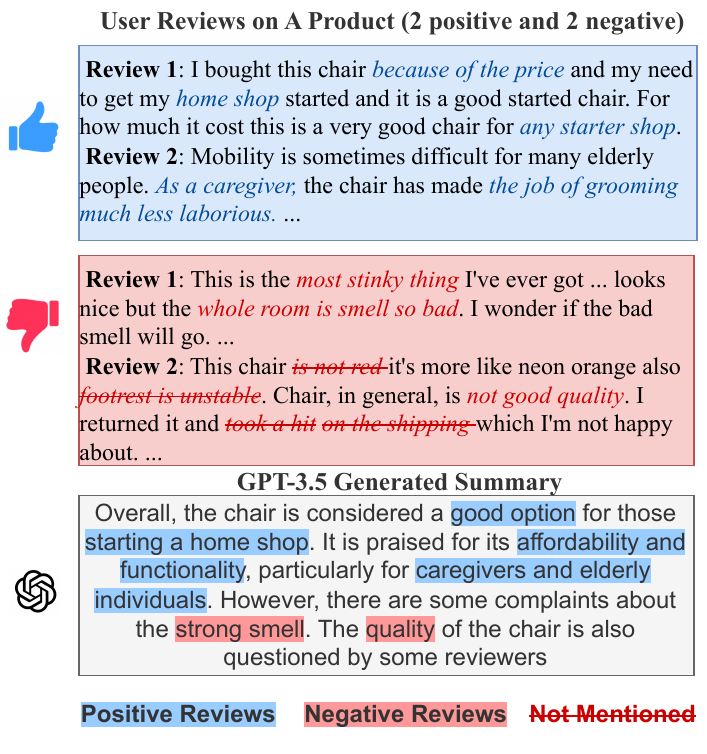}
\caption{An example from \fairdata. The blue/red box displays the input consisting of positive/negative reviews. The grey box shows the summary generated by GPT-3.5 (text-davinci-003). The generated summary is unfair because the negative reviews are underrepresented compared with the positive reviews.}
\label{fig:example}
\end{figure}

\section{Introduction}
Different social and demographic groups of people hold diverse and potentially even conflicting perspectives and opinions, which are expressed in user-generated text data in various domains and topics such as product reviews, social debates, healthcare, law, and politics~\cite{shandilya2018fairness, brazinskas2020few, zhang-etal-2016-conversational, huang2023examining, huang2023embrace, kovavc2023large, hayati2023far}.
When a summarization system is faced with diverse perspectives that can be equally correct and fundamental, a fair summary shall provide an accurate and comprehensive view of diverse perspectives from these groups.
For example, Figure~\ref{fig:example} shows the reviews about an Amazon product. The summary generated by GPT-3.5~\citep{Ouyang2022instructgpt} unfairly represents sentiments because it summarizes more content from positive reviews while ignoring some critical points from negative reviews.

However, existing summarization metrics cannot adequately measure fairness. Metrics such as ROUGE~\citep{lin-2004-rouge} and BERTScore~\citep{bert-score} fail to measure fairness for two reasons. First, they are not inherently designed for quantifying fairness, but rather measuring similarity between system outputs and references based on n-gram overlapping or embeddings. Second, reference summaries can also be biased. Moreover, current work on summarization~\citep{shandilya2018fairness} and LLM evaluation~\citep{chang2023survey} has not explored the fairness of abstractive summarization, which is more prevalent than extractive summarization in the era of LLMs~\cite{openai2023gpt4}.

In this work, we present the first study that systematically \textit{defines}, \textit{quantifies}, and \textit{benchmarks} the fairness of abstractive summarization using large language models in a wide range of domains and topics. 
First, we define our task of fair abstractive summarization as generating summaries that are fair with respect to a social attribute that can take different values, such as the sentiment attribute with positive/negative values.
We curate a benchmark \fairdata (Table~\ref{tab:fairsumm_data}) by unifying and cleaning six existing datasets in domains where fairness is a critical issue including healthcare, politics, restaurant, product, law, and policy, which covers social attributes of gender, party, sentiment, rating, and speaker. The fairness of the summary on these attributes greatly influences the information and even the opinions of the readers. 
Next, we quantify fairness in abstractive summarization as a distribution alignment between the generated summary and the source text. The distribution is calculated by ratios of semantic units of different social attribute values. According to the definition, we propose several metrics to evaluate fairness, including Binary Unfair Rate (BUR) which determines unfair summaries by checking if any social attribute value is under-represented,
Unfair Error Rate (UER) which measures the distance between the generated summary and the source text in terms of their social attribute value distributions.
Third, we benchmark fairness in abstractive summarization by using \fairdata to conduct a comprehensive evaluation of nine main-stream LLMs, including GPT (text-davinci-003~\citep{Brown2020gpt3}, gpt-turbo-3.5~\citep{Ouyang2022instructgpt}, gpt-4~\citep{openai2023gpt4}), LLaMA (Alpaca~\citep{alpaca}, llama2-chat-7/13/70B~\citep{touvron2023llama}), PaLM 2~\citep{anil2023palm} (text-bison@001), and Claude~\citep{claude} (claude-instant-1). 

Results of our metrics and human evaluation show that neither human-written reference summaries nor LLM-generated summaries could maintain fairness based on our definition. Even GPT-4 suffers from fairness problems as the BUR score of GPT-4 on Amazon reviews reaches 74.27\%. 
We also provide three simple but effective ways to improve the summary fairness including changing decoding temperature, changing summary length, and appending the definition of fairness to the instruction prompt. The result shows that BUR decreases from 51.11\% to 47.48\% on gpt-turbo-3.5 by providing instructions of fairness definition.

Our contributions are as follows: (1) We propose new metrics for quantifying and measuring the fairness of abstractive summarization over diverse perspectives. (2) We collect and unify a comprehensive benchmark for analyzing fairness of abstractive summarization in various domains and topics. (3) We conduct comprehensive experiments to investigate the fairness of reference summaries and summaries generated by popular LLMs, finding that both reference/generated summaries often fail to achieve fairness. (4) We analyze the factors that influence fairness to propose three simple but effective methods to improve fairness.

\section{Summarization of Diverse Perspectives with Social Attributes} 
In this section, we first formulate our task of summarization from diverse perspectives with social attributes (Section~\ref{sec:task}), then present data collection of \fairdata for this task (Section~\ref{sec:data}).

\subsection{Task Formulation}
\label{sec:task}

\paragraph{Social Attributes.}
A fair abstractive summary should include comprehensive perspectives without underrepresenting certain groups of people.
We use social attributes to indicate the properties that form groups of people. As shown in Table~\ref{tab:fairsumm_data}, social attributes can be gender, party, and sentiment.

\phantomsection \label{def:2.1}\paragraph{Definition 2.1 Summarization with Social Attribute.} A summarization dataset $\mathcal{D}$ consists of pairs of source text and target summary $(\bm{x}, \bm{y})$. Here, $\bm{x}=[x_1,x_2,\cdots,x_n]$ is the source text consisting of $n$ token $x_i$ and $\bm{y}=[y_1,y_2,\cdots,y_m]$ is the target summary consisting of $m$ tokens $y_i$. We consider a social \textit{attribute} $a$ having $r$ different values $\mathcal{V} = \{v_1, v_2, \cdots, v_r\}$, $r>1$. Furthermore, we use $v(\cdot)$ to denote the value of a token. We assume each source unit $x_i$ belongs to one value $v(x_i) \subset \mathcal{V}$, $|v(x_i)|=1$, while each token in target is associated with one or multiple values $v(y_i) \subseteq \mathcal{V}$, $|v(y_i)|\geq 1$.
For instance, when considering gender as an attribute $a$ in Twitter review summarization, the $r=2$ possible values are $\mathcal{V}=\{\texttt{male}, \texttt{female}\}$. For each tweet token $x_i$ in the source, if it is written by a male, $v(x_i) = \texttt{male}$, otherwise $v(x_i) = \texttt{female}$\footnote{To simplify the task setting, we consider two genders in our experiments, while our approach can generalize.}. For each token $y_i$ in the summary of tweets, it could come from $\texttt{male}$, $\texttt{female}$, or both. For instance, if both groups claim a product is ``good''. Then ``good'' can be both from $\texttt{male}$, and $\texttt{female}$.

\begin{table*}[t!]
\resizebox{\textwidth}{!}{
\begin{tabular}{@{}lllllrrr@{}}
\toprule
Dataset& Domain & Source Form  & Collect From  & Attributes& Values & \# Samples & Max and Avg Length\\ \midrule
Claritin  & Healthcare & Social media & \url{twitter.com} & Gender& 2  & 1350  & 1087/572.2  \\
US Election& Politics & Social media & \url{twitter.com}& Party & 3  & 1350  & 1247/677.7\\
Yelp& Restaurant  & Review& \url{yelp.com}& Sentiment& 3 & 1500  & 576/402.1\\
Amazon & Product & Review& \url{amazon.com} & Rating& 5& 1500  & 531/346.5 \\
Supreme Court & Law & Dialogue & \url{supremecourt.gov} & Speakers & 3-10  & 665& 2721/1910.9  \\
Intelligence Squared (IQ2) & Public Policy & Debate& \url{opendebate.org}  & Speakers & 2-10& 1421  & 3275/1650.7\\ \bottomrule
\end{tabular}
}
\caption{Dataset characteristics of \fairdata for fair abstractive summarization. Examples in Table~\ref{tab:dataset_example}.
}
\label{tab:fairsumm_data}
\end{table*}

\section{Definition of Fairness in Summarization}
\label{sec:fair}
In this section, we provide definitions for fairness. Figure~\ref{fig:metrics} shows the logic of computation of fairness in abstractive summarization and corresponding evaluation metrics. The first stage is to produce the value distribution. Given a sample pair of source and target text, we first split the source according to the annotated values of social attributes. Then, we apply N-gram/BERT/BART scores to compare target tokens with source tokens to obtain the value of each token in the summary. Next, we count the number of tokens for each value in the source and target and obtain the value distribution of them. 

\subsection{Defining Value Distribution}
Computation of source and target value distribution (source/target distribution for short) is the first step of fairness computation. The formal definition is listed as follows. 

\paragraph{Definition 3.1 Value Distribution.} We define the value distribution $\bm{p}$ in three scenarios: (1) the value distribution in source: $\bm{p}_{\bm{x}} = [\bm{p}_{\bm{x}}(v_1), \bm{p}_{\bm{x}}(v_2),\cdots,\bm{p}_{\bm{x}}(v_r)]$, where $\bm{p}_{\bm{x}}(v_k)$ is the proportion of source tokens associated with value $v_k$, $0\leq \bm{p}_{\bm{x}}(v_k) \leq 1$, $\sum_{k=0}^r \bm{p}_{\bm{x}}(v_k) = 1$. (2) Similarly, the value distribution in target: $\bm{p}_{\bm{y}} = [\bm{p}_{\bm{y}}(v_1), \bm{p}_{\bm{y}}(v_2),\cdots,\bm{p}_{\bm{y}}(v_r)]$. (3) a gold value
distribution $\bm{p}_{\bm{g}} = [\bm{p}_{\bm{g}}(v_1),\bm{p}_{\bm{g}}(v_2), \cdots, \bm{p}_{\bm{g}}(v_r)]$ which means the distribution that aligned with user's requirement (Section~\ref{sec:define_fair}). 

\subsection{Calculating Value Distribution}
To compute $\bm{p}_{\bm{x}}$, since the meta-data of the dataset shows the values of each token in the source, it can be acquired by counting the number of tokens of each value. However, the calculation of $\bm{p}_{\bm{y}}$ cannot be obtained easily due to the abstract nature of summaries.
As abstractive summarization does not directly copy from the source text, it contains words that are not in the source but still semantically come from the source.
This makes it difficult to attribute a sentence in the summary to a certain part of the source. 
To tackle this, we propose two algorithms for calculating the target distribution $\bm{p}_{\bm{y}}$:

\paragraph{N-gram Matching.}
For a given k, we scan each k-gram in target text, if this k-gram is also shown in the source text, we assign this k-gram the value of the source text it appears. It is worth noting that the k-gram in the target can be assigned multiple values according to the definition. These metrics can automatically compute $\bm{p}_{\bm{y}}$ given $\bm{p}_{\bm{x}}$, with no additional annotation on $\bm{p}_{\bm{y}}$. We use $k=1$ for all experiments. We use N-gramScore to represent this matching approach. 

\paragraph{Neural Matching.}
We use BARTScore~\citep{yuan2021bartscore}, and BERTScore~\cite{bert-score} to measure the distance between the target and source. 
These metrics capture the similarity in semantics, overcoming the new word matching issues in the n-gram algorithm~\cite{banerjee-lavie-2005-meteor}. Specifically, for a given source $\bm{x}$ and target $\bm{y}$, we first split $\bm{x}$ by value. For each $v_i \in \mathcal{V}$, $\bm{x_i} = \{x \in \bm{x} | v(x) = v_i\}$. Let $r = |\mathcal{V}|$. Then, we compute the correlation between $\bm{x_i}$ and $\bm{y}$ to obtain $\bm{p}_{\bm{y}}$: $\bm{p}_{\bm{y}} = \text{Softmax}(\text{Score}(\bm{x_1},\bm{y}),\cdots, \text{Score}(\bm{x_{r}}, \bm{y}))$,
where $\text{Score}$ is BARTScore and BERTScore, and the temperature of $\text{Softmax}$ is a hyperparameter. We pick 0.1 for experiments because it better aligns with human evaluation (Appendix~\ref{sec:meta-eval}).

The design of the metrics is different in terms of the following two aspects. First, they have different granularity. N-gramScore is based on the token which controls the fairness of the token level. BERTScore computes the similarity of the sentences which is in sentence level. BARTScore computes via the entailment of the entire summary. This is on the summary level. Second, these scores have different advantages. N-gramScore can be applied to diverse length source text while the other two scores have length limitations. BERT score can capture semantic similarity while BART score captures entailment. These two models rely on the accuracy of the backbone models as well. 
\subsection{Defining Summarization Fairness}
\label{sec:define_fair}
After defining the source/target distribution, the goal of fair summarization is equivalent to producing a $\bm{p}_{\bm{y}}$ not underrepresenting any value in $\bm{p}_{\bm{x}}$.
\paragraph{Definition 3.2 Summarization Fairness.}
We define an unfair summary as exhibiting an underrepresentation of user groups in summaries. Given value distributions, we can define several types of fairness: (1) \textbf{Ratio Fairness}: the target value distribution $\bm{p}_{\bm{y}}$ shall follow the same value distribution as source $\bm{p}_{\bm{x}}$. In this case, $\bm{p}_{\bm{g}} = \bm{p}_{\bm{x}}$. (2) \textbf{Equal Fairness}: the target value distribution $\bm{p}_{\bm{y}}$ shall follow the uniform value distribution $\bm{p}_{\bm{g}} = [1/r,1/r,\cdots,1/r]$, regardless of source. This distribution keeps the balance of each value. (3) Furthermore, users can define any $\bm{p}_{\bm{g}}$ to indicate their ideal distribution, not restricted to Ratio Fairness or Equal Fairness. Without loss of generality, we will discuss ratio fairness in our metric and experiment sections, because summarization aims to compress but still follow the original distribution in the source text~\cite{shandilya2018fairness}.

\section{\fairdata}
\label{sec:data}
To conduct a comprehensive analysis of fairness, we select and then process existing datasets to form our fair summarization benchmark \fairdata.
We follow two main principles to choose six datasets from a large variety of existing datasets. First is the \textit{quality}, the source texts need to be human-written and 
marked with clear, and precise social attributes. These attributes are existing metadata in the datasets, except for the sentiment attribute, which is derived from classifier predictions. Second is \textit{diversity}, the collected datasets need to cover various domains and perspectives. 
Table~\ref{tab:fairsumm_data} shows the statistics of \fairdata. It covers six different domains with six attributes. 
Overview of these six datasets are listed as follows (details in Appendix~\ref{sec:data_construction}): 

\textbf{Claritin}~\citep{shandilya2018fairness} contains tweets about the effects of the drug Claritin where $a=\text{gender}$ and $\mathcal{V}=\{\text{male},\text{female}\}$. \textbf{US Election}~\citep{shandilya2018fairness} contains tweets posted during the 2016 US Presidential election where $a=\text{politics}$ and $\mathcal{V}=\{\text{pro-rep},\text{pro-dem}, \text{neu}\}$, meaning pro-republic, pro-democracy, or neutral. For Claritin and US-Election, we randomly sample tweets of different values with a certain ratio and then combine them as one input source. 
\textbf{Amazon} and \textbf{Yelp} are two datasets from the FewSum dataset~\citep{brazinskas2020few} containing product and business reviews. 
Each sample in these two datasets consists of eight user-written reviews. In Amazon dataset, $a=\text{ratings}$, and $\mathcal{V}=\{1,2,3,4,5\}$ shows the ratings of each reviewer. 
In Yelp dataset, $a=\text{sentiment}$, and $\mathcal{V}=\{\text{pos},\text{neu},\text{neg}\}$ displays the sentiment of reviews. 
\textbf{SupremeCourt}~\citep{danescu2012echoes} contains a collection of conversations from the U.S. Supreme Court Oral Arguments, where $a=\text{speakers}$, $\mathcal{V}$ are the names of all participants. \textbf{Intelligence Squared Debate dataset (IQ2)}~\citep{zhang-etal-2016-conversational} collects public debates that follow the Oxford style and are recorded live, where $a=\text{speakers}$, $\mathcal{V}$ contains the names of all participants. For SupremeCourt and IQ2, we truncate each transcript of the debate into shorter segments of $k$ tokens. We include the whole text of each sample for the experiments of claude-100k.
\begin{figure}[t!]
\centering
\includegraphics[width=\linewidth]{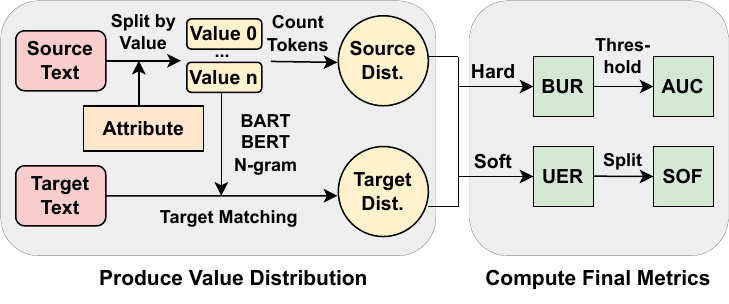}
\caption{Overview of our proposed metrics. Dist. means value distribution.}
\label{fig:metrics}
\end{figure}
\section{Evaluating Fairness in Summarization}
As shown in Figure~\ref{fig:metrics}, we define four types of metrics to evaluate whether the generated summaries are fair: Binary Unfair Rate (BUR), Unfair Error Rate (UER), Area Under Curve (AUC), and Second-Order Fairness (SOF).

\paragraph{Definition 5.1 Binary Unfair Rate (BUR).} We define Binary Unfair Rate (BUR), a binary function that outputs 1 if the sample
is unfair; and 0 otherwise. To be specific, we define the summary as fair if and only if $\bm{p}_{\bm{y}}(v_k) \geq
\tau \cdot \bm{p}_{\bm{x}}(v_k)$ holds for each value $k = 1, 2, \cdots, r$, where $0 \leq \tau \leq 1$ is a tolerance hyperparameter that is chosen based on application scenarios~\citep{shandilya2018fairness}. Otherwise, if $\bm{p}_{\bm{y}}(v_k) <
\tau \cdot \bm{p}_{\bm{x}}(v_k)$, we say the value $v_k$ is underrepresented. Thus
\begin{equation}
\small
f_{\text{BUR}}(\bm{x},\bm{y};\tau) = \mathbbm{1}\left(\bigvee_{i=1}^r \bm{p}_{\bm{y}}(v_k) < \tau \cdot \bm{p}_{\bm{\bm{x}}}(v_k)\right) 
\end{equation}

This metric is either zero to one and it can be averaged over the dataset, demonstrating the proportion of summaries that are \textit{unfair} in the dataset. However, this hard metric may not be able to describe which one is more unfair/fair if BUR of two samples is the same. Thus, we propose Unfair Error Rate as a soft metric to compute the sum of the distance to source distribution of each value.
\paragraph{Definition 5.2 Unfair Error Rate (UER).}
This metric defines a function $f_{\text{UER}}(\bm{x},\bm{y})$ measuring the distance between $\bm{p}_{\bm{x}}$ and $\bm{p}_{\bm{y}}$. UER serves as the supplementary metric for BUR as it shows more fined-grained details compared with BUR.
\begin{equation}
f_{\text{UER}}(\bm{x},\bm{y}) = \frac{1}{r}\sum_{k=1}^r \max(0, \bm{p}_{\bm{x}}(v_k) -\bm{p}_{\bm{y}}(v_k) )
\end{equation}

UER shows the average percentage of values that are underrepresented.  Furthermore, since people have different tolerance and requirements for fairness, we propose Area Under Curve to compute the expected result of BUR when the tolerance is randomly picked from zero to one. 

\paragraph{Definition 5.3 Area Under Curve (AUC).}
As mentioned, $\tau$ is a hyperparameter chosen according to the scenarios. However, people may have different assignments of $\tau$ according to their own experiences. To incorporate different tolerance hyper-parameters, we propose AUC of fairness threshold (AUC for short) for BUR metric $f_{\text{AUC}}(\bm{x},\bm{y})$.
\begin{equation}
f_{\text{AUC}}(\bm{x},\bm{y}) = \int_{0}^{1} f_{\text{BUR}}(\bm{x},\bm{y};\tau) \; d\tau 
\end{equation}

It is a number between zero and one showing the expected BUR before assigning a fairness threshold $\tau$ to it because we assume that the users evenly pick the threshold between zero and one. It is worth noting that, practically, we sample ten points evenly to approximate the integration.

These three metrics cannot show which value is more unfair in one sample. Thus, we further propose Second-Order Fairness to compute the variance of UER among all values.
\paragraph{Definition 5.4 Second-Order Fairness (SOF).} 
Second-Order Fairness (SOF) $f_{\text{SOF}}$ measures the unfairness difference among all values, defined as:
\begin{equation}
\begin{split}
& S_\text{UER} =\{\max(0, \bm{p}_{\bm{x}}(v_k) -\bm{p}_{\bm{y}}(v_k) )\ | k \in [r]\} \\
& f_{\text{SOF}}(\bm{x},\bm{y}) = \frac{1}{r}\sum_{s \in S_\text{UER}} \left|s- \frac{\sum_{s \in S_\text{UER}}s}{r}\right|
\end{split}
\end{equation}
Inspired by~\citet{macqueen1967some}, SOF measures the coherence of UER set $S_{\text{UER}}$ via computing the average distance to the center $\sum_{s \in S_\text{UER}}s/r$.
If all values' UER are close to their averaged center, the possibilities of being unfair are similar for values. In this case, each value has a similar possibility to be unfair, so-called second-order fairness.

\paragraph{Tolerance Hyperparameter.} Regarding $\tau$, the threshold that decides whether a value is under-represented is based on the application scenarios~\citep{celis2018fair}. When strict fairness is required, such as the political opinions of two parties, this threshold should be set to a high value, for instance, $\tau = 0.8$ or higher~\citep{shandilya2018fairness}. However, in some other text summarization tasks in our daily lives, we do not require each value to have strict constraints, such as product review summarization. In the extreme case, we only want each value to appear in the summary rather than constraining the time of appearance, so we can set $\tau = 0$. In this paper, we set $\tau = 0.8$ for all experiments~\citep{biddle2017adverse}, if not specified.

\paragraph{Metric Quality Validation.} We validate our metrics because they correlate with human evaluation (Appendix~\ref{sec:meta-eval}) and achieve lower/upper bound values on extreme synthetic examples (Appendix~\ref{sec:metric_bound}).

\begin{table*}[!ht]
\centering
\resizebox{\textwidth}{!}{
\begin{tabular}{@{}lrrrrrrrrrrrrr@{}}
\toprule
&\multicolumn{1}{c}{} & \multicolumn{2}{c}{Claritin}& \multicolumn{2}{c}{US Election}& \multicolumn{2}{c}{Yelp}& \multicolumn{2}{c}{Amazon}  & \multicolumn{2}{c}{SupremeCourt}  & \multicolumn{2}{c}{IQ2} \\ \cmidrule(lr){3-4} \cmidrule(lr){5-6} \cmidrule(lr){7-8} \cmidrule(lr){9-10} \cmidrule(lr){11-12} \cmidrule(lr){13-14}
&\multicolumn{1}{c}{Size}& \multicolumn{1}{c}{BUR$\downarrow$} & \multicolumn{1}{c}{UER$\downarrow$} & \multicolumn{1}{c}{BUR$\downarrow$} & \multicolumn{1}{c}{UER$\downarrow$} & \multicolumn{1}{c}{BUR$\downarrow$} & \multicolumn{1}{c}{UER$\downarrow$} & \multicolumn{1}{c}{BUR$\downarrow$} & \multicolumn{1}{c}{UER$\downarrow$} & \multicolumn{1}{c}{BUR$\downarrow$} & \multicolumn{1}{c}{UER$\downarrow$} & \multicolumn{1}{c}{BUR$\downarrow$} & \multicolumn{1}{c}{UER$\downarrow$} \\ \midrule
Alpaca & 7B   & 65.38 & 10.35 & 87.33 & 8.40  & 44.04 & 5.11  & 80.73 & 7.16  & 97.99 & 8.79  & 91.06 & 8.99 \\ 
llama-2-chat & 7B   & 62.99 & {\ul 9.09}& \textbf{78.64}   & 6.69  & 41.76 & 3.59  & 76.00 & \textbf{5.09}& 97.40 & 4.92  & \textbf{84.41}   & \textbf{6.02} \\ 
llama-2-chat & 13B  & 63.06 & 9.35  & {\ul 79.28}  & 6.79  & \textbf{36.20}   & {\ul 3.53}& \textbf{73.51}   & {\ul 5.11}& \textbf{95.49}   & {\ul 4.58}& 86.04 & 6.66  \\ 
llama-2-chat & 70B  & \textbf{61.53}   & 9.92  & 79.78 & \textbf{6.41}& {\ul 37.04}  & \textbf{2.81}& {\ul 74.69}  & 5.63  & 96.94 & 4.74  & 85.71 & 6.64 \\ \midrule
text-bison@001 & N/A  & 67.73 & 10.49 & 86.53 & 8.71  & 45.65 & 5.37  & 84.42 & 8.43  & 97.74 & 5.65  & 89.26 & 8.21 \\ 
text-davinci-003 & 175B & {\ul 62.94}  & \textbf{9.08}& 82.74 & 7.09  & 43.09 & 4.35  & 79.89 & 6.60  & 97.39 & 5.25  & 87.17 & 7.20  \\ 
gpt-turbo-3.5& 175B & 64.30 & 9.18  & 81.38 & {\ul 6.48}& 38.64 & 4.00  & 75.89 & 5.82  & {\ul 96.59}  & 4.64  & {\ul 84.52}  & {\ul 6.07}\\ 
gpt-4 & N/A  & 66.37 & 9.94  & 79.93 & 6.99  & 39.42 & 3.72  & 74.78 & 5.49  & 96.79 & \textbf{4.57}& 87.71 & 6.94  \\ \bottomrule 
\end{tabular}}
\caption{Main results with our proposed metrics. We report the mean of N-gramScore, BERTScore, and BARTScore. Full results are in Table~\ref{tab:overall_detail}.
BUR and UER are better with a lower score $\downarrow$. \textbf{Bold} is the best performance, and \underline{underline} is the second best. 
}
\label{tab:overall}
\end{table*}

\begin{table*}[t!]
\begin{minipage}[t][][b]{0.43\linewidth}
\centering
\resizebox{.8\textwidth}{!}{
\begin{tabular}{@{}lrr@{}}
\toprule
 & \multicolumn{2}{c}{N-gramScore} \\ \cmidrule(lr){2-3}
 & \multicolumn{1}{c}{BUR$\downarrow$} & \multicolumn{1}{c}{UER$\downarrow$}   \\ \midrule
\textit{\textbf{claude-instant-1 (N/A)}} & & \\
SupremeCourt & 99.07  & 1.77\\
IQ2 & 100.00 & 2.37\\ \bottomrule
\end{tabular}
}
\caption{Long context results. Only N-gramScore is used because the input length exceeds the limit of BERTScore and BARTScore.}
\label{tab:cluade}
\end{minipage}
\hspace{2mm}
\begin{minipage}[t][][b]{0.55\linewidth}
\centering
\resizebox{.95\textwidth}{!}{
\begin{tabular}{@{}lrrrrrr@{}}
\toprule
 & \multicolumn{2}{c}{N-gramScore}  & \multicolumn{2}{c}{BERTScore}  & \multicolumn{2}{c}{BARTScore}  \\ \cmidrule(lr){2-3} \cmidrule(lr){4-5} \cmidrule(lr){6-7} 
 & \multicolumn{1}{l}{BUR$\downarrow$} & \multicolumn{1}{l}{UER$\downarrow$} & \multicolumn{1}{l}{BUR$\downarrow$} & \multicolumn{1}{l}{UER$\downarrow$} & \multicolumn{1}{l}{BUR$\downarrow$} & \multicolumn{1}{l}{UER$\downarrow$} \\ \hline
\textbf{\textit{reference summary}} & \multicolumn{1}{l}{}& \multicolumn{1}{l}{}& \multicolumn{1}{l}{}& \multicolumn{1}{l}{}& \multicolumn{1}{l}{}& \multicolumn{1}{l}{}\\
Amazon& 95.00 & 18.50 & 95.00 & 18.50 & 91.67 & 9.44  \\
Yelp & 68.00 & 26.43 & 53.04 & 7.90  & 56.00 & 10.11 \\ \hline
\textbf{\textit{gpt-turbo-3.5}}& \multicolumn{1}{l}{}& \multicolumn{1}{l}{}& \multicolumn{1}{l}{}& \multicolumn{1}{l}{}& \multicolumn{1}{l}{}& \multicolumn{1}{l}{}\\
Amazon& 68.33 & 3.97  & 76.67 & 5.99  & 93.33 & 9.51  \\
Yelp & 21.00 & 3.12  & 36.00 & 4.76  & 64.00 & 6.27  \\ \bottomrule
\end{tabular}
}
\caption{Comparison between human-written reference summary and summary generated by gpt-turbo-3.5.}
\label{tab:gold}
\end{minipage}
\end{table*}

\section{Experiment Results and Analysis}
Our results and analysis aim to answer the following research questions:
\begin{itemize}[noitemsep,topsep=0pt,parsep=0pt,partopsep=0pt]
\item RQ 1: How fair are the summaries generated by LLMs based on our metrics (Section~\ref{sec:overall_result})?
\item RQ 2: How fair are the existing human-written reference summaries according to our automatic metrics (Section~\ref{sec:reference_summary})?
\item RQ 3: How do humans perceive the fairness of summaries generated by LLMs (Section~\ref{sec:human_evaluation})?
\item RQ 4: How can we dissect the fairness of summaries generated by LLMs using SOF (Section~\ref{sec:second_order_fairness}) and AUC (Section~\ref{sec:AUC})?
\end{itemize}
We test nine LLMs as listed in Appendix~\ref{sec:model}. For analysis, we use gpt-turbo-3.5 on the Claritin dataset with N-gramScore as the target matching method, if not specified. 

\subsection{Overall Results}
\label{sec:overall_result}
Table~\ref{tab:overall} shows the results of all models using our metrics.
In general, \textit{our results indicate that many summaries generated by LLMs are not fair}. Most models do not perform well according to our metrics. For example, on Amazon datasets, gpt-4 can achieve a BUR score of 75\%. This means around 3 out of 4 summaries of reviews have some bias towards one side or the other. Overall, LLaMAs enjoy better fairness than other models. We found gpt-turbo-3.5 and gpt-4 are in general better than their older version text-davince-003, and 13B/70B llama2-chat are better than their smaller 7B version and Alpaca. However, we don’t find strong evidence that gpt-4 is better than gpt-turbo-3.5.
Besides, we observe that the performance of gpt-4 and gpt-turbo-3.5 significantly vary on different male ratios and source lengths (Appendix~\ref{sec:model_comparison}).
\paragraph{Long Context Results.}
Different from other models, claude-instant-1 can consume 100k tokens as inputs. Thus, we prepare a full dataset on IQ2 and SupremeCourt without segmentation. As shown in Table~\ref{tab:cluade}, compared with short input, it is more difficult to maintain a fair summary. Thus, \textit{Claude also suffers from fairness issues on long input datasets}.

\subsection{Fairness of Reference Summary}
\label{sec:reference_summary}
We also explore the fairness of the human-written reference summaries. We analyze the fairness of 100 reference summaries compared with summaries generated by gpt-turbo-3.5 on Amazon and Yelp datasets. As shown in Table~\ref{tab:gold}, the reference summary also surfers from high BUR and UER scores, indicating that \textit{existing human-written reference summaries cannot maintain the fairness either, even worse than those generated by gpt-turbo-3.5}. 
This echoes previous findings that LLMs can generate summaries with higher quality than human-written ones~\cite{goyal2022zeroshotnews}. 

\begin{table*}[!t]
\centering
\resizebox{0.83\textwidth}{!}{
\begin{tabular}{@{}llllllllll@{}}
\toprule
& \multicolumn{3}{c}{Claritin} & \multicolumn{3}{c}{Yelp}& \multicolumn{3}{c}{IQ2} \\ \cmidrule(lr){2-4} \cmidrule(lr){5-7} \cmidrule(lr){8-10}
& \multicolumn{1}{c}{BUR$\downarrow$} & \multicolumn{1}{c}{UER$\downarrow$} & \multicolumn{1}{c}{Rating$\downarrow$} & \multicolumn{1}{c}{BUR$\downarrow$} & \multicolumn{1}{c}{UER$\downarrow$} & \multicolumn{1}{c}{Rating$\downarrow$} & \multicolumn{1}{c}{BUR$\downarrow$} & \multicolumn{1}{c}{UER$\downarrow$} & \multicolumn{1}{c}{Rating$\downarrow$} \\ \midrule
Alpaca  & 76.47& 13.10& 85.00  & 45.00& 4.67& 88.24  & 88.88& 11.83& 100.00 \\ \midrule
text-davinci-003 & \textbf{58.75}& \textbf{8.85} & 51.76  & \textbf{15.00}& \textbf{2.58} & 64.71  & \textbf{60.00}& 10.90& 100.00 \\
gpt-turbo-3.5& 82.35& 10.72& \textbf{45.00}& 42.11& 3.33& \textbf{62.50}& 80.00& \textbf{8.39} & 100.00 \\
gpt-4& 76.17& 13.07& 80.00  & 40.00& 2.76& 88.24  & 100.00& 15.74& 100.00 \\ \bottomrule
\end{tabular}}
\caption{Human evaluation results. BUR and UER are from the first stage. Rating is from the second stage.}
\label{tab:human1}
\end{table*}

\subsection{Human Evaluation}
\label{sec:human_evaluation}

We present two sets of human evaluation scores, corresponding to the two stages of the annotation process (more design details in Appendix~\ref{sec:design_human}).
In the first stage, we first use a tool provided by ~\citet{liu2022revisiting} to decompose each sentence into Atomic Content Units (ACUs). Then, we ask the annotators to verify the ACUs and then count the proportions of each value in the summary. We collect the results and compute BUR and UER scores.
In a pilot annotation, the inter-annotator agreement of this stage is 68.97\% Krippendorff’s alpha, showing high agreement. 
As shown in Table~\ref{tab:human1}, text-davinci-003 achieves the best performance on all the metrics in ACU-based scores, showing that human favors text-davinci-003 the most. It is worth noting that, Alpaca achieves fairness that is similar to that of gpt-turbo-3.5, demonstrating that small models can achieve similar fairness.  

In the second stage, the annotators are asked to give a rating on the fairness of the summaries. This is more subjective because annotators can put different emphasis on different ACUs. This serves as another angle of fairness which gives more space to annotators' personal judgment on fairness. 
The Randolph's Kappa score is 0.41, demonstrating the high agreement across 4 annotators. As shown in Table~\ref{tab:human1}, different from the ACU-based scores, \textit{the overall ratings lean toward turbo models which align with the automatic metrics}.

\subsection{Meta-Evaluation}
\label{sec:meta-eval}
\begin{figure}[!t]
\centering
\includegraphics[width=\linewidth]{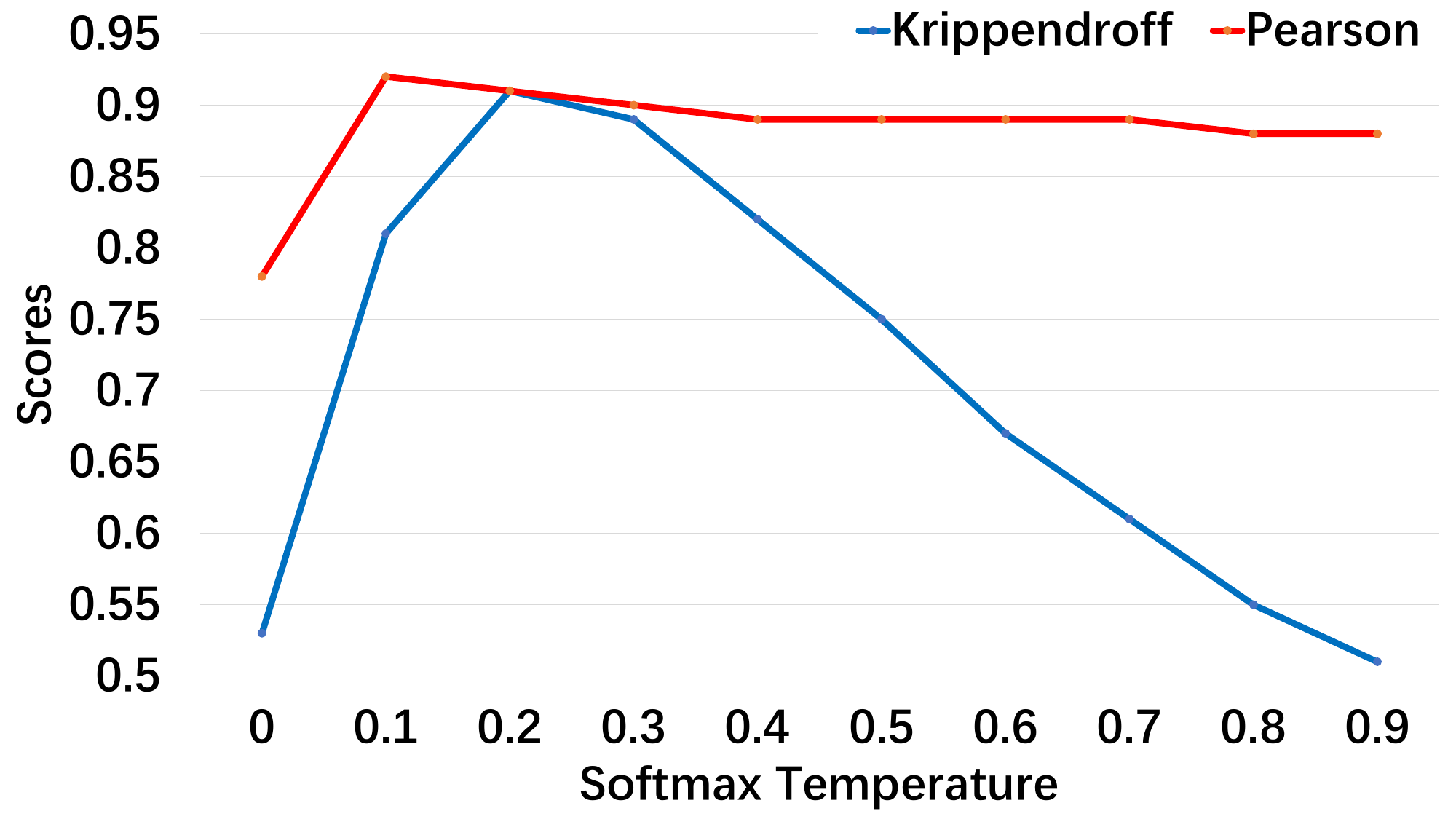}
\caption{Relation between temperature and correlation scores on Claritin using gpt-turbo-3.5. X-axis is the softmax temperature of BARTScore. Y-axis is the Krippendorff’s alpha and Pearson correlation coefficient with human evaluation.
Pearson correlation coefficient is higher than Krippendorff's alpha because Pearson correlation coefficient only computes positive relations while Krippendorff's alpha requires the annotations to be the same.} 
\label{fig:temp}
\end{figure}

We also use human evaluation to test the quality of proposed metrics. As shown in Figure~\ref{fig:temp}, the correlation between the target distribution computed by BARTScore and human evaluation is around 0.91 when choosing 0.1 as the softmax temperature. This shows the high alignment of proposed metrics and human evaluation. Because BARTScore is not designed to show the proportion of entailment (For instance, 0.5 does not indicate that half of the summary can be inferred from the source.), we aligned BARTScore with human evaluation on absolute values by selecting 0.1 as temperature which enjoys a high Krippendorff’s alpha with human evaluation.

\subsection{Distribution across Values}
\label{sec:second_order_fairness}
\begin{figure}[!t]
\centering
\includegraphics[width=\linewidth]{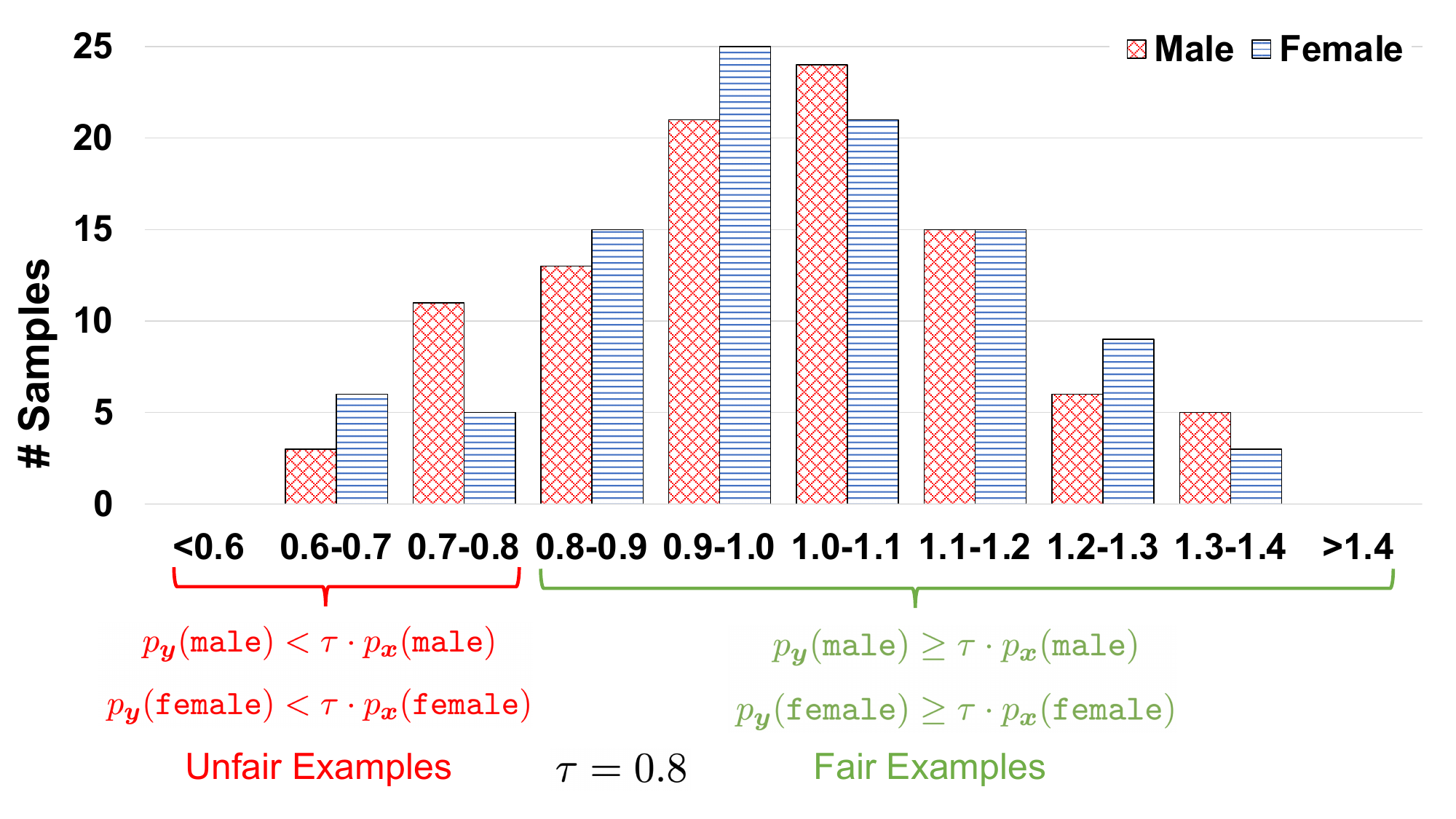}
\caption{Distribution of Male and Female values in summaries generated by gpt-turbo-3.5 on Claritin.}
\label{fig:value_dis}
\end{figure}

\begin{table}[t!]
\centering
\resizebox{\linewidth}{!}{
\begin{tabular}{@{}lccc@{}}
\toprule
  & N-gram$\downarrow$ & BERT$\downarrow$ & BART$\downarrow$ \\ \midrule
\textbf{SupremeCourt} &&\\
Alpaca	& 37.04 &	47.72 &	42.24 \\ 
llama-2-chat-7b	& 25.14	& 36.48	& 41.75 \\ 
llama-2-chat-13b&	25.34&	\textbf{31.59}&	\textbf{40.72} \\ 
llama-2-chat-70b&	\textbf{24.46}	&34.23	&\underline{41.19} \\ 
\hline
text-bison@001&	27.81&	35.55&	42.11 \\
text-davinci-003 & 25.83 & 34.49 & 42.19 \\
gpt-turbo-3.5& 25.40 & \underline{32.33} & 41.68 \\
gpt-4 & \underline{24.86} & 32.36 & 41.48 \\
claude-instant-1  & 40.29 & -- & -- \\ 
\hline
\textbf{IQ2}  &&&\\
Alpaca	&29.94	&43.86	&48.29 \\
llama-2-chat-7b	&\underline{25.59}	&40.76	&47.66 \\
llama-2-chat-13b	&26.34	&\textbf{40.13}	&\textbf{46.85} \\
llama-2-chat-70b	&25.63	&49.06	&47.80 \\
\hline
text-bison@001	&32.05	&43.22	&\underline{47.13} \\
text-davinci-003 & 27.32 & 41.97 & 48.42 \\
gpt-turbo-3.5 & \textbf{24.00} & \underline{40.24} & 47.68 \\
gpt-4 & 31.26 & 42.60 & 48.46 \\
claude-instant-1  & 33.47 & -- & --\\ \bottomrule
\end{tabular}}
\caption{AUC scores on SupremeCourt and IQ2. 
}
\label{tab:AUC}
\end{table}

We dissect the fairness of abstractive summaries generated by LLMs using the distribution across values. Figure~\ref{fig:value_dis} shows the distribution of males and females on Claritin. X-axis is the target distribution divided by source distribution $\bm{p}_{\bm{y}}(v_k)/\bm{p}_{\bm{x}}(v_k)$. As can be seen, about 20\% of gpt-turbo-3.5 outputs contain underrepresented and unfair summaries for both female and male contents with the tolerance hyperparameter $\tau = 0.8$. \textit{The proportion of unfair male/female tweets is similar, showing that the model generally performs well on Second Order Fairness (SOF is 0.17).}

\subsection{Analysis on AUC}
\label{sec:AUC}
We compute the AUC score by aggregating BUR scores over different cutoff thresholds. 
This is useful for datasets containing more values $|\mathcal{V}|$ that require more flexible fairness. Thus, we compute the score for SupremeCourt and IQ2 datasets in Table~\ref{tab:AUC}. As shown in the table, LLaMA 2 obtains the best AUC score on SupremeCourt and IQ2. \textit{This demonstrates that across different thresholds, LLaMA 2 obtains better performance than the GPT family on datasets with diverse values.} Claude model obtains a higher AUC score compared with other models as it consumes much longer data.

\section{Improving Fairness of Abstractive Summarization}
We propose three methods to improve the fairness of summaries by LLMs: decoding temperature, summary length, and fairness instruction.
\subsection{Improving through Hyperparameters}
\begin{figure}[!t]
\centering
\includegraphics[width=\linewidth]{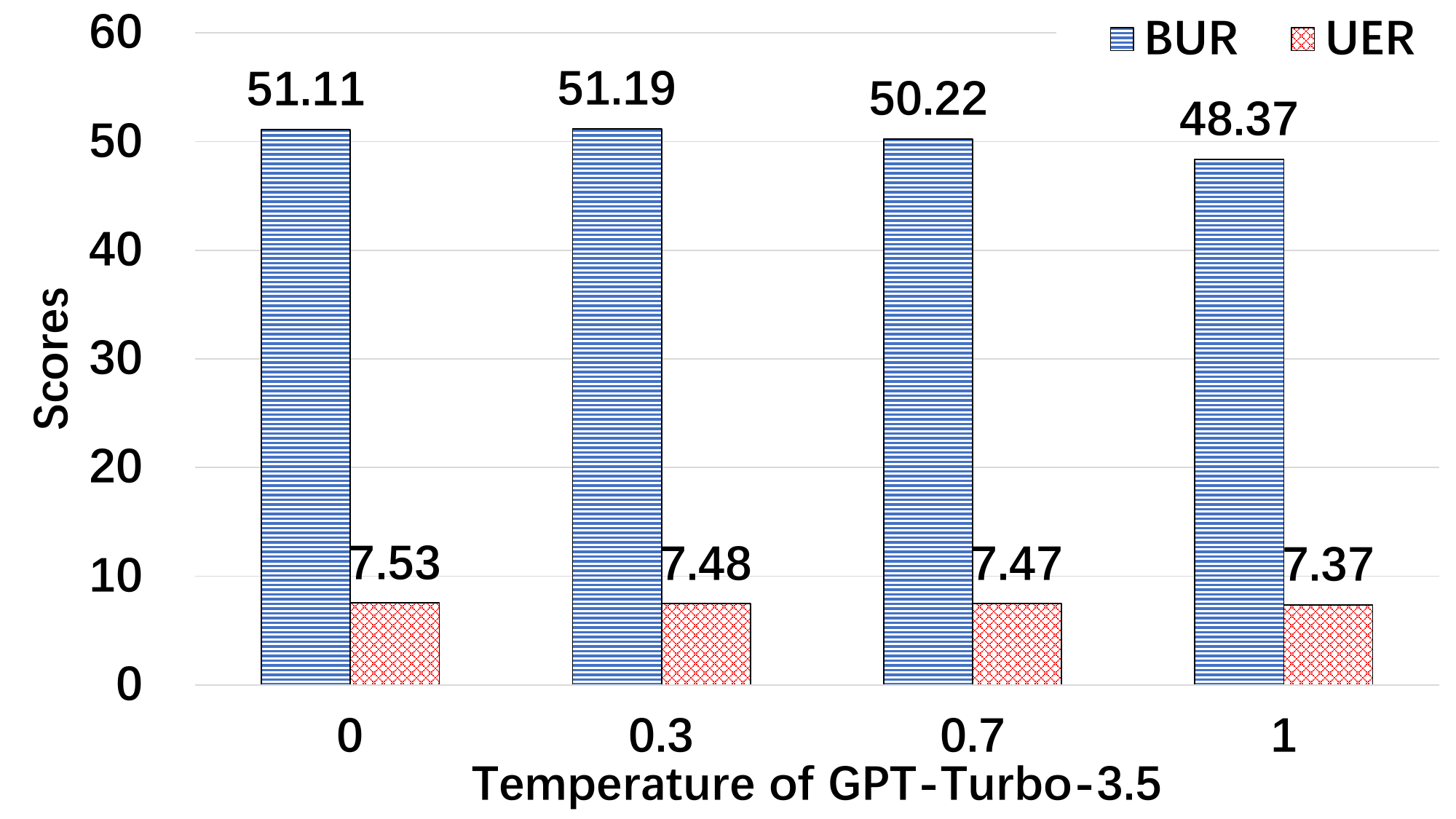}
\caption{Effect of decoding temperature. 
}
\label{fig:model_temp}
\end{figure}

\paragraph{Decoding Temperature.}
The temperature controls the proportion of novel words in outputs by modifying the sampling probability of tokens~\citep{hinton2015distilling}. The model generates more novel words with higher temperatures (Appendix~\ref{sec:example}). To test the effect of temperature towards fairness, we evaluate gpt-turbo-3.5 on various temperatures from $\{0,0.3,0.7,1\}$ on Claritin. As shown in Figure~\ref{fig:model_temp}, \textit{the BUR and UER scores decrease significantly when the temperature rises}. This shows the model generates more diverse results when the temperature is higher, leading to more fairness in certain samples. However, the readability of the summary will be lower as some noise tokens are generated with higher temperatures. Thus, it is a trade-off between quality and fairness.

\begin{figure}[!t]
\centering
\includegraphics[width=\linewidth]{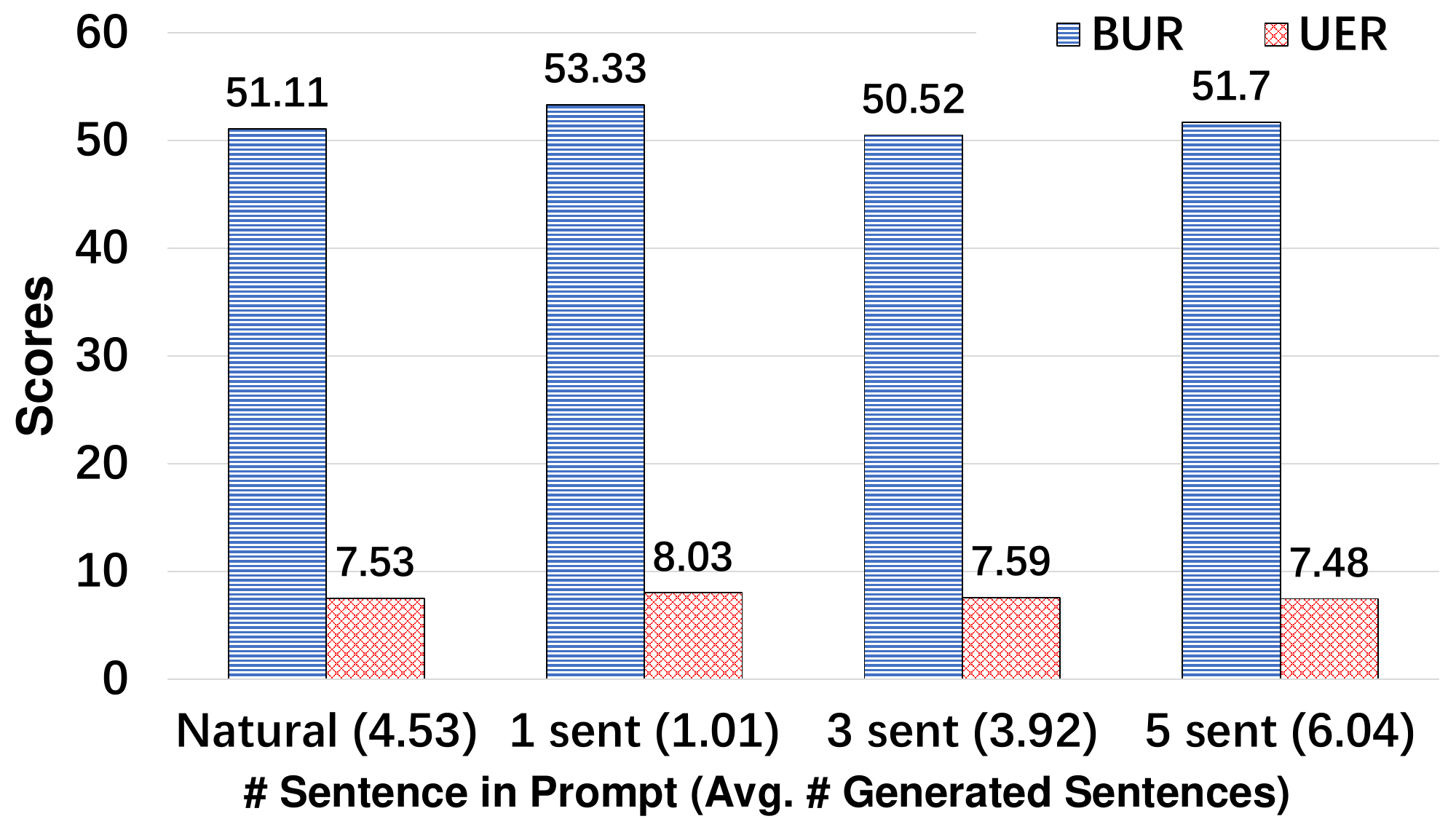}
\caption{Effect of summary length in the number of sentences. Natural means not controlling the number of sentences, while 1/3/5 sent means that the model is prompted to generate a summary with 1/3/5 sentences.
}
\label{fig:sent}
\end{figure}

\paragraph{Summary Length.}
We analyze the influence of the length by using gpt-turbo-3.5 on Claritin to generate different numbers of sentences in summaries (Table~\ref{tab:template}). Figure~\ref{fig:sent} shows the BUR and UER scores for different lengths. Among them, 3 sentences demonstrate the best performance on fairness, while the 1 sentence shows the worst. \textit{Therefore, medium length works best, and when there are too many/fewer sentences, balancing the value in summary contents is more difficult.}


\subsection{Improving through Instructions}
We modify the instruction by giving more information and definition of fairness in the prompt (Appendix~\ref{sec:template}). Table~\ref{tab:strategies} shows the performance of gpt-3.5-turbo on three datasets. 
\textit{We found that adding the definition of fairness summaries can significantly improve fairness}. However, SOF increases after using definition instruction which shows that the models are biased towards one gender. 
\begin{table}[t!]
\centering
\resizebox{0.9\linewidth}{!}{
\begin{tabular}{@{}llll@{}}
\toprule
 & BUR$\downarrow$ & UER$\downarrow$& SOF$\downarrow$\\ \midrule
Claritin & 51.11 & 7.53 & \textbf{0.17} \\
\;\; + Definition & \textbf{47.48} & \textbf{7.18} & 0.98 \\ \midrule
Yelp & 21.80 & 3.01 & \textbf{1.25} \\
\;\; + Definition  & \textbf{21.73} & \textbf{2.98} & 1.27 \\ \midrule
Election & 72.67 & 5.00 & \textbf{0.54} \\
\;\; + Definition  & \textbf{67.04} & \textbf{4.35} & 1.03 \\ \bottomrule
\end{tabular}}
\caption{Effect of fairness instruction prompt. 
Adding the definition of fairness can significantly improve fairness.}
\label{tab:strategies}
\end{table}

\section{Related Work}
\paragraph{Fairness and Bias in Natural Language Generation.}
People have been aware of social fairness and bias in natural language generation such as gender bias in machine translation and toxicity from prompt-based generation~\cite{prates2020assessing,sheng2020towards,gehman2020realtoxicityprompts}.
Only a few efforts study the fairness of summarization~\cite{carenini2008extractive,shandilya2018fairness,jorgensen2021evaluation,zhou2023characterizing,chhikara2023fairness,liu2023constitutes}. \citet{10.1145/3359274} proposed an analysis of whether the generated summaries fairly represent different social groups, such as gender or politics. \citet{keswani2021dialect} proposed a framework that takes an existing text summarization algorithm as a black box to obtain a summary that is relatively more dialect-diverse. However, they are limited to unsupervised extractive approaches and restricted domains. 
Another relevant line of research is opinion summarization~\cite{carenini2013multi,gerani2014abstractive,bravzinskas2019unsupervised,bravzinskas2020few,bhaskar2022zero,iso2021comparative}, yet existing work mainly focused on few-shot learning instead of fairness as collecting reference summaries can be expensive.
Therefore, our goal is to develop new metrics and datasets for fair abstractive summarization over diverse perspectives and values. \citet{feng-etal-2023-pretraining} and \citet{santurkar2023whose} examined political biases in LMs stemming from training data and model persona, whereas our proposed benchmark assesses the ability of LLMs to generate fair abstractive summarizations.

\paragraph{Summarization Evaluation.}
Recent research reveals that the prior metrics such as ROUGE~\citep{lin-2004-rouge} and its variations~\citep{rankel-etal-2013-decade} cannot reliably evaluate LLM-generated summaries~\citep{fabbri2021summeval,pillutla2021mauve,tam2022evaluating,goyal2022news,zhang2023benchmarking,chang2023booookscore}. 
New metrics of other dimensions have been proposed such as factuality~\cite{kryscinski2019evaluating} and controllability~\cite{zhang2022macsum}.
\citet{liu2022revisiting, liu2023towards} propose a two-stage metric that predicts the Atom Content Unit (ACU) for the summary first and checks the proportion of the ACUs that can be inferred from the source text. \citet{olabisi-etal-2022-analyzing} proposed a benchmark for measuring the ability to represent salient as well as diverse perspectives. However, their evaluation metrics are unable to assess the distribution of summaries in abstractive summarization tasks.

\paragraph{Evaluation of LLMs.}
Recently, more attention has been attracted to the evaluation of LLMs~\citep{hendrycks2020measuring,liang2022holistic,srivastava2022beyond,workshop2022bloom,li2023alpacaeval,zheng2023judging,chen-etal-2023-unisumm}. 
Some work evaluates the zero-shot ability~\citep{qin2023chatgpt}, and multitasking, multilingual, and multimodal abilities~\citep{bang2023multitask}, while others focus on the alignment and safety issues~\cite{perez2022red,wang2023decodingtrust,schulhoff2023ignore}.
By contrast, we probe the fairness of LLMs by benchmarking fair abstractive summarization.

\section{Conclusion and Future Work}
In this paper, we systematically investigate fair abstractive summarization. 
Results of our metrics and human evaluation show that neither humans nor LLMs could maintain the fairness of summaries. Further analysis shows that prompt engineering and careful choosing of the temperature of LLMs can significantly improve the performance of fairness.

Future work includes fine-tuning small or large language models to improve the fairness of abstractive summarization, extending metrics to other types of bias, and developing datasets that apply the concept of fairness in other generation tasks.

\section*{Acknowledgment}
We thank Greg Durrett, Shuyang Cao, Ranran Haoran Zhang, Sarkar Snigdha Sarathi Das, Fang Geng, Xi Li, and Kai Huang for the valuable discussions and comments. We also would like to thank the anonymous reviewers for their helpful comments.
\clearpage
\newpage

\section*{Limitations}
Our scope of the project is to focus on model fairness instead of bias in the dataset. We assume that the fairness of the training data of the models is unknown and only discuss the fairness of the generated summarization results. 
We believe we have covered a variety of domains and attributes in many socially impactful applications where fairness is critical, while our framework of evaluation can also be extended to other domains and attributes. In this paper, all datasets in \fairdata are pre-cleaned by their respective authors, so we assume there are no invalid samples. However, we acknowledge the possibility of invalid samples.

\section*{Ethics Statement}
\fairdata contains six datasets that cover the typical domains and attributes that could cause fairness issues. We do not mean that these domains are the only ones that need attention for fairness. 
Due to the characteristics of the experiment, we pick some of the diverse domains and attributes to form \fairdata. Second, we define the values for each attribute in a convenient and efficient way. It is possible that some values do not cover all possibilities, such as classifying gender in the Claritin dataset into female and male. 
Again, we use this dataset as the basic material for experiments, not meaning the source of \fairdata is exactly aligned with social norms.
Third, as indicated in the paper, our definition of fairness is flexible, and not restricted to ratio fairness. We conduct experiments using this criterion because it is one of the definitions that can reflect the fairness issue and enhance our conclusions. It does not show that ratio fairness is more important than other ones. In fact, our definition and code support various types of fairness, including but not restricted to ratio fairness, and equal fairness. 
Forth,
we use the results to measure the fairness of the models, but this does not mean that our scores are the only metrics to judge the fairness of the summaries. Last but not least, the formation of \fairdata may not align with real-world distribution due to the preprocessing algorithm. For Claritin, and Election, we randomly sample a certain number of twits to form one sample, however, these twits may not be able to reflect the real-world distribution. For Yelp and Amazon, we randomly sampled eight reviews which may also be different from real-world data. Also, for sentiment in the Yelp dataset, we use the NLTK tool to predict the sentiment of each review which may lead to misclassifications, though NLTK sentiment classification can achieve a high score. For dialogue datasets IQ2 and Oxford Debates, we segment the source into chunks due to the lengthy input. This may hurt the integrity of origin dialogue, leading to a difference between \fairdata and original complete dialogue in these datasets.

\bibliography{anthology,custom}
\bibliographystyle{acl_natbib}

\appendix
\section{Dataset Construction}
\label{sec:data_construction}
Details of these six datasets are listed as follows:

\textbf{Claritin}~\citep{shandilya2018fairness} contains tweets about the effects of the drug Claritin.
Each tweet is annotated with the gender of the user who posted it. We use this dataset to evaluate gender fairness, where $a=\text{gender}$ and $\mathcal{V}=\{\text{male},\text{female}\}$. We randomly sample a number of tweets written by males and females with a certain ratio between them and then combine them together as one input source. This procedure is repeated multiple times with a diverse number of tweets and male ratios to form the whole dataset.

\textbf{US Election}~\citep{shandilya2018fairness} contains tweets posted during the 2016 US Presidential election. Each tweet is annotated as supporting or attacking one of the presidential candidates or neutral or attacking both. We use this dataset to evaluate politics fairness across parties, where $a=\text{politics}$ and $\mathcal{V}=\{\text{pro-rep},\text{pro-dem}, \text{neu}\}$. Similar to the Claritin dataset, we form the final dataset as a mixture of sampled tweets of certain ratios of each value in $\mathcal{V}$.

\textbf{Amazon} and \textbf{Yelp} are two datasets from the FewSum dataset~\citep{brazinskas2020few} containing product and business reviews. 
Each sample in these two datasets consists of multiple user-written reviews. 
Among these two datasets, FewSum provides reference summaries for 32 products on Amazon and 70 businesses on Yelp. 
We use the Amazon dataset to conduct opinion fairness analysis, where $a=\text{ratings}$, and $\mathcal{V}=\{1,2,3,4,5\}$ shows the ratings of each reviewer. 
We use the Yelp dataset to conduct sentiment fairness analysis, where $a=\text{sentiment}$, and $\mathcal{V}=\{\text{pos},\text{neu},\text{neg}\}$ displays the sentiment of reviews. 
We use NLTK~\cite{loper2002nltk} to produce sentiment for source text in Yelp because the dataset does not provide the sentiment of source reviews. We also use the reference summaries of the two datasets to further analyze the fairness of human-written summaries (Section~\ref{sec:reference_summary}).

\textbf{SupremeCourt}~\citep{danescu2012echoes} contains a collection of conversations from the U.S. Supreme Court Oral Arguments
from 204 cases involving 11 Justices and 311 other participants. We use SupremeCourt to analyze the fairness of speakers in court transcripts, where $a=\text{speakers}$, $\mathcal{V}$ are the names of all participants. 
For the input to the models, we truncate each transcript into shorter segments of $k$ tokens to ensure the texts do not exceed the length limit of the models. We also include the whole text of each sample for the experiments of claude-instant-1.

\textbf{Intelligence Squared Debate dataset (IQ2)}~\citep{zhang-etal-2016-conversational} collects public debates that follow the Oxford style and are recorded live. In each debate, teams of 2 to 3 experts debate for or against a motion and attempt to sway the audience to take their position. Each debate also has a moderator. We use the IQ2 dataset to analyze the fairness of speakers in debates, where $a=\text{speakers}$, $\mathcal{V}$ contains the names of all participants. We also truncate each transcript of the debate into shorter segments of $k$ tokens. We include the whole text of each sample for the experiments of claude-100k.

It is worth noting that, two dialogue datasets contain more than one attribute, which emphasizes that people can have diverse perspectives of fairness on the same input source text.

\section{Dataset Example}
\label{sec:dataset_example}
Table~\ref{tab:dataset_example} shows some examples in \fairdata.
\begin{table*}[!t]
\centering
\small
\begin{tabular}{lp{13cm}}
\Xhline{5\arrayrulewidth}
  Claritin  &  @dararoseee have you seen the 1st one? word, i thought i'd avoided them, but i had a sneezing fit earlier this week. claritin on deck. || Ima have to dope myself up on dis Zyrtec, Benadryl, r Claritin. || Butttttt I have some cute things, Claritin, a tank full of gas. So I guess I'll survive. || @brooke\_opland there my babies omg can't live without my Claritin ... \textit{truncated contents} ... @nancyholtzman i heard that allergy medicine (antihistamine) can also dry up milk, depending on strength, claritin ok but.. @Jeannette108 || It has taken at least 3 hours for the Claritin to kick in and make my nose slow from a running faucet, to just a leaky one. || @thebrokenplate i just hope it IS just allergies.  Is your throat sore too? Took claritin last nt but now have to wait 24 hrs to take again || Dudes, I think the Claritin is making me...peculiar. I mean, moreso than usual.",\\
  \midrule
  Election & Very strong case for Hillary by the Times, with context and perspective that's been hard to find this race.  https://t.co/n8mJPZEjEf || Gold Star families have made sacrifices most of us can't even fathom. They deserve our respect and our thanks. https://t.co/c9gDHudBjt || Michael Bloomberg knows Donald Trump. And he's begging us to elect Hillary Clinton. Trump is dangerous. \#ImWithHer https://t.co/JWC3rAb8Td || Happy birthday to this future president. https://t.co/JT3HiBjYdj ... \textit{truncated contents} ... || This election is a choice between an economy that benefits everyone or an economy that benefits...Donald Trump. https://t.co/PEHnJDdiLq || We can take on the threat of climate change and make America a clean energy superpower.  Or, we can do nothing. https://t.co/JlYmN61epB || Weird, I can't find anything on @Project\_Veritas videos on @nytimes or @washingtonpost. It's almost like they're deliberately ignoring them.",\\
  \midrule
  Yelp & I was pleasantly surprised with my meal . The burger and fries were seasoned to perfection . The willow bar dessert was mouth watering . The prices weren 't bad and the staff was friendly and helpful . We will be back || Stopped there just for a drink and a little something to eat . We loved the wine list and the food . The menu had interesting selections and everything was really beautifully prepared and delicious . Worth a trip . || Fantastic food . We had the Butternut squash ravs and hubby had the pork chop . Amazing . Drinks were great . Waitstaff cool and in the game . Can 't go wrong here ! ! ... \textit{truncated contents} ... || Easter dinner did not disappoint . The filet , and crab cakes are fantastic . Only wish we had bread served with dinner . Great job ! Service was great as usual . || Great Dining experience . 730 reservations on a Saturday night . Seated promptly , waiter was attentive and informative . It was our first visit here so we followed his suggestions and it was spot on . Food was very good and atmosphere was great . A very good dining experience .", \\ \midrule
  Amazon & I know that the point of these bags is for them to be for small snack items , but they were too small for most of what I wanted to use them for . I think the normal sandwich sized bags work just as well . || I used these for travel sized items being I hate spillage in my suitcase . I store small items such as loose herbs in them for immediate use . ... \textit{truncated contents} ... || THESE BAGS ARE THE BES , WE USE THEN IN EVERY WAY POSSIBLE AT HOME AND WHEN THE KIDS HAVE TO GO TO SCHOOL . || I love snack size bags and am happy to find them on Amazon . These bags are perfect for packing snacks on the go without using bigger sandwich bags . I also use these smaller bags to organize office and hair supplies . || I got 6 total boxes . I love the ziploc ones because of the double seal . The bags are reusable for about 5 times before they get nasty .", \\ \midrule
  SupremeCourt & \textbf{JUSTICE STEVENS} : We will now hear argument in the Cherokee Nation against Thompson and Thompson against the Cherokee Nation. Mr. Miller. \textbf{MR. MILLER} : Justice Stevens, and may it please the Court: These two contract cases concern whether the Government is liable in money damages under the Contract Disputes Act and section 110 of the Indian Self-Determination Act when the Secretary fails to fully pay a contract price for the -- \textbf{JUSTICE O'CONNOR} : Would you mind explaining to us how these two cases relate? The Court of Appeals for the Federal Circuit decision went one way and the Tenth Circuit went another. And are the claims at all overlapping? How are they differentiated? \textbf{MR. MILLER} : No, Justice O'Connor. They're -- they're not overlapping. T ... \textit{truncated contents} ... \textbf{MR. MILLER} : Your Honor, we -- we do not believe that that -- that should be the outcome. That would advantage the contractors that came forward and not take account of the entire situation. We think the global situation has to be looked at. The total amount of the contracts that were not paid in fiscal year 1994 –\\ \midrule
  IQ2 & \textbf{Bob Costas} : … And now I’d like to introduce Robert Rosenkranz, who is the chairman of the Rosenkranz Foundation, and the sponsor of Intelligence Squared, who will frame tonight’s debate. Bob? This is Bob. \textbf{Bob Costas} : Thank you again, Bob. So this is the sixth debate of the second Intelligence Squared US Series. ... \textit{truncated contents} ... \textbf{Robert Rosenkranz} : Well thank you very much. And, uh, uh, on behalf of, uh, Dana Wolfe, our executive producer and myself, uh, I’m just, uh, thrilled to welcome you. When we scheduled this event some, uh, five months ago, we had no idea it would be so timely. Just in the past month the, uh, Mitchell Report was released, naming some eighty eight Major League Baseball players alleged to have used steroids and, uh, uh, other drugs. Roger Clemens’ denials have been heard in 60 Minutes and were front page news in Sunday’s New York Times. Uh, record breaking sprinter Marion Jones was sentenced to six weeks in prison-, or six months, I ... \textit{truncated contents} ... \\

\Xhline{5\arrayrulewidth}
\end{tabular}
\caption{Six samples from \fairdata. One sample for each dataset. Reviews are shuffled and concatenated by ``||''. The names of the speakers in SupremeCourt and IQ2 datasets are in \textbf{bold}.}
\label{tab:dataset_example}
\end{table*}

\section{Quality of the Metrics}
\label{sec:metric_bound}
We conduct an oracle test to evaluate our metric quality. To this end, we create extreme synthetic examples by biased/balanced sampling to test the upper/lower bound of BUR and UER scores. We randomly choose 100 examples from Claritin, and we replace the original summary with some tweets sampled from the source as the pseudo-summary. For biased sampling, we sample 5 male tweets from the source as the pseudo-summary, which should have high BUR and UER scores. For balanced sampling, we sample 5 tweets from the source as the pseudo-summary so that the target distribution is the same as the source, e.g., when the male ratio is 20\%, we sample 1 male tweet and 4 female tweets from 10 source tweets to maintain the proportion. Although this is not strictly fair, because the length of each tweet varies, it should enjoy a very low BUR and UER scores.

\begin{figure}[t!]
\centering
\includegraphics[width=\linewidth]{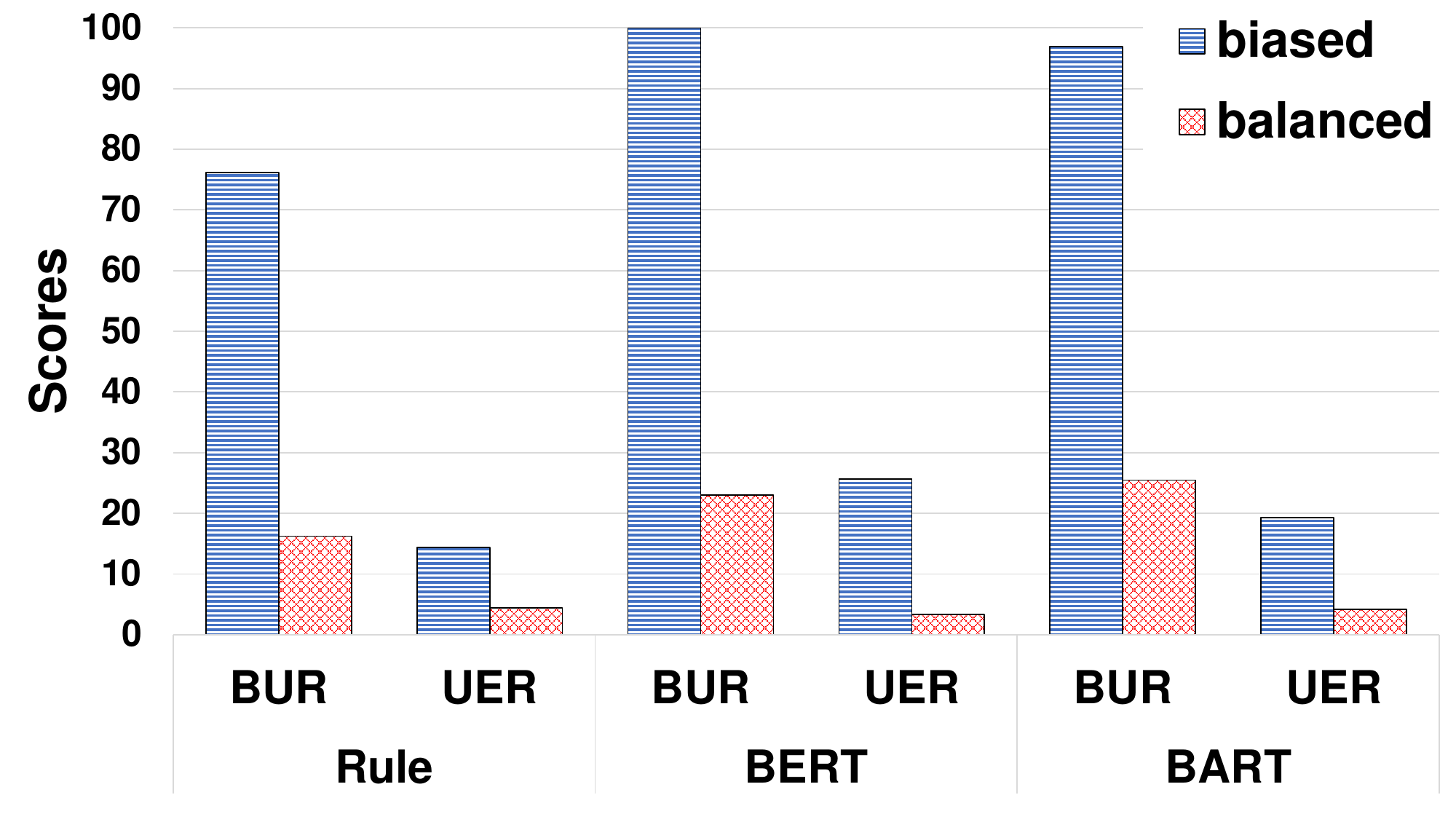}
\caption{Extreme synthetic cases to validate the quality of metrics.}
\label{fig:extreme}
\end{figure}

As shown in Figure~\ref{fig:extreme}, The difference is significant between scores of bias/balanced samplings for all three metrics. Among the three scores, N-gramScore gives the lowest BUR/UER scores in biased sampling. This is because N-gramScore decomposes the summary of sampled male tweets into tokens during matching. Although the entire summary comes from male tweets, many tokens appear in female tweets in the source. This decreases the BUR/UER. In contrast, the other two metrics are based on semantic similarities.

Besides, we pick $k=1$ for N-gramScore because using $k > 1$ makes the metrics meaningless. During our experimentation with higher values of $k$ ($k=2, k=3$), we observed that in over 80\% of samples, the overlap between the source and target summaries was zero. This is because our abstractive summarization tasks, particularly for shorter summarizations, rarely produce 2-gram or 3-gram overlaps. Consequently, using $k > 1$ makes the metrics meaningless as the fairness of the generated summary will equal generating an empty string. This underscores the necessity for our embedding-based metrics in such scenarios.

\section{Correlation between Metrics}
\begin{table}[]
\centering
\begin{tabular}{@{}lllll@{}}
\toprule
& BUR  & UER  & AUC  & SOF  \\ \midrule
BUR & 1.00 & 0.56 & 0.44 & 0.08 \\
UER & 0.56 & 1.00 & 0.27 & 0.43 \\
AUC & 0.44 & 0.27 & 1.00 & 0.06 \\
SOF & 0.08 & 0.43 & 0.06 & 1.00 \\ \bottomrule
\end{tabular}
\caption{Pearson correlation between metrics on Claritin dataset.}
\label{tab:metrics_correlation}
\end{table}
We conduct an analysis to calculate the correlation between metrics, and the results are summarized in Table~\ref{tab:metrics_correlation}. As can be seen, BUR has a high correlation with UER and AUC which is in line with our assertion: BUR provides a binary indication of fairness, while UER and AUC offer detailed information across different UER set $S_{\text{UER}}$ or thresholds $\tau$, respectively. Similarly, the strong correlation between SOF and UER supports our design rationale, as UER represents the mean of the UER set, while SOF reflects the coherence of this set. When UER is high, it suggests that the elements in the UER set are likely dispersed along the axis.

\section{Human Evaluation}
We introduce the design principles of human evaluation, meta-evaluation, and evaluation guideline details.
\label{sec:design_human}
\subsection{Design of Human Evaluation} 

To clearly measure the performance of the models and validate our automatic metrics, 
we design a two-stage human evaluation by asking four annotators with English expertise and experience in summarization to manually label 20 examples in Claritin, 20 in Yelp, and 10 in IQ2.

The two annotation tasks include Sentence Fact Identification and Summary Fairness Identification. Sentence Fact Identification produces a score using step-by-step instruction because the annotators need to read and identify facts and add them together to produce target value distribution. On the contrary, Summary Fairness Identification gives an end-to-end score that directly labels the rating of fairness after reading the definitions. We would like to use these two types of explicit and implicit computation of fairness to make human evaluation more comprehensive and diverse.

\paragraph{ACU-based Score.} 
The first task is Sentence Fact Identification. This task asks annotators to label the target value distribution of the summaries. Annotators are firstly asked to split the summary sentence into facts where each fact is the atom semantic unit which is defined as the smallest unit that holds semantic information. Next, the source text is split into different values, and the annotators need to identify where each atom semantic unit comes from, which means the input perspectives that can fully infer (entail) this fact. Compared with direct operation on each sentence, this operation enjoys two advantages. First, each fact contains the same amount of information so that they can be added together without defining weights. Second, it is easier for the annotator to judge where it is from.  
After obtaining the human-annotated target distribution based on the ratio of atom semantic units, we run the automatic program to compute the BUR and UER metrics. We call this an ACU-based human evaluation score. 

\paragraph{Rating-based Score.} The second task is Summary Fairness Identification. In this stage, the annotators are asked to judge the fairness of each value. Different from stage one's computation, annotators need to give an overall judgment of the level of fairness.
We assign 5 ratings, from 1 to 5, where 1 indicates only representing 0\%-20\% of the input distribution and 5 indicates over 80\%. Thus, only 5 indicates a fair summary, while 1-4 is all unfair with different levels. 
We call this a rating-based human evaluation score. More details of annotation guidelines are available in Appendix~\ref{sec:human_evaluation_details}.

\subsection{Guidelines of Human Evaluation}
\label{sec:human_evaluation_details}
We list part of the guidelines for annotators, including instructions for Sentence Fact Identification and Summary Fairness Identification. We also show our annotation interface.

\begin{figure*}
\centering
\includegraphics[width=\textwidth]{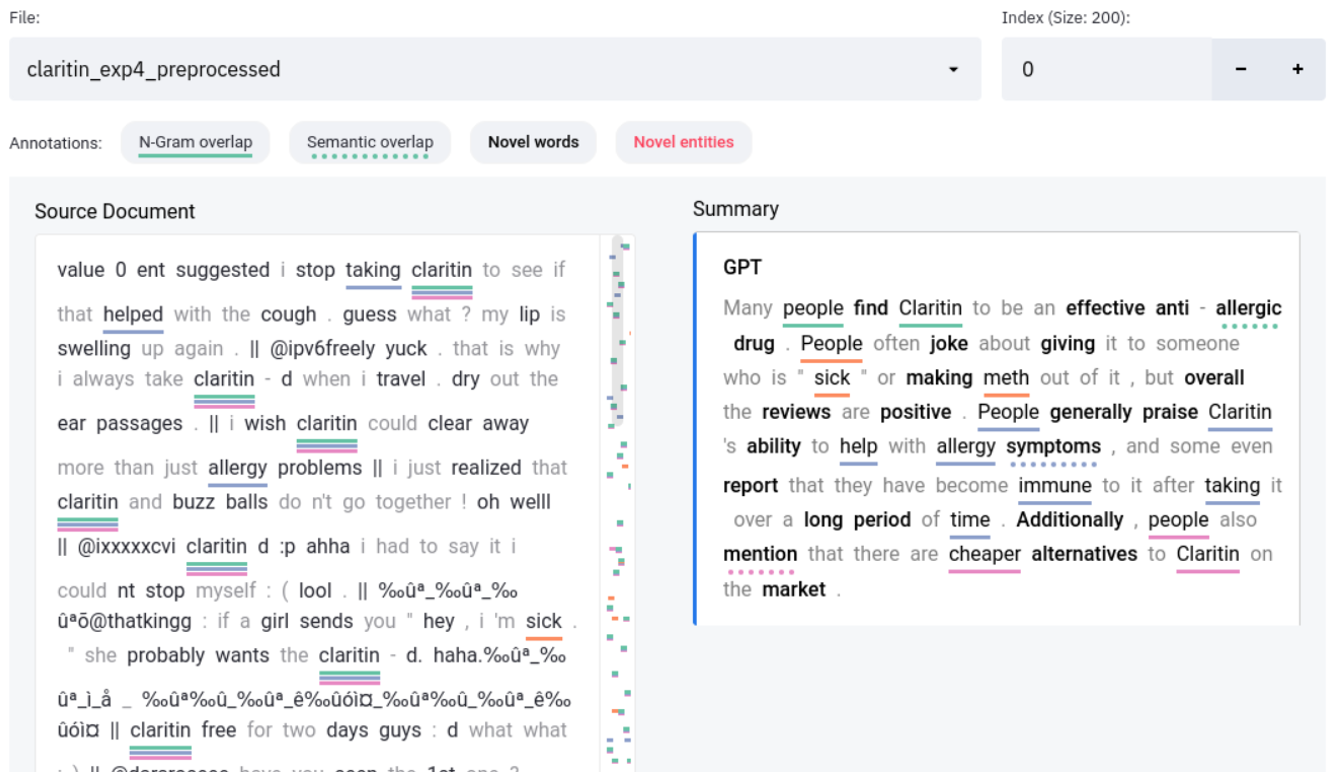}
\caption{A sample for SummVis annotation interface. There are 4 types of tokens labeled by SummVis: N-gram Overlap, Semantic Overlap, Novel words, and Novel entities. }
\label{fig:summvis}
\end{figure*}

\begin{figure*}
\centering
\includegraphics[width=\textwidth]{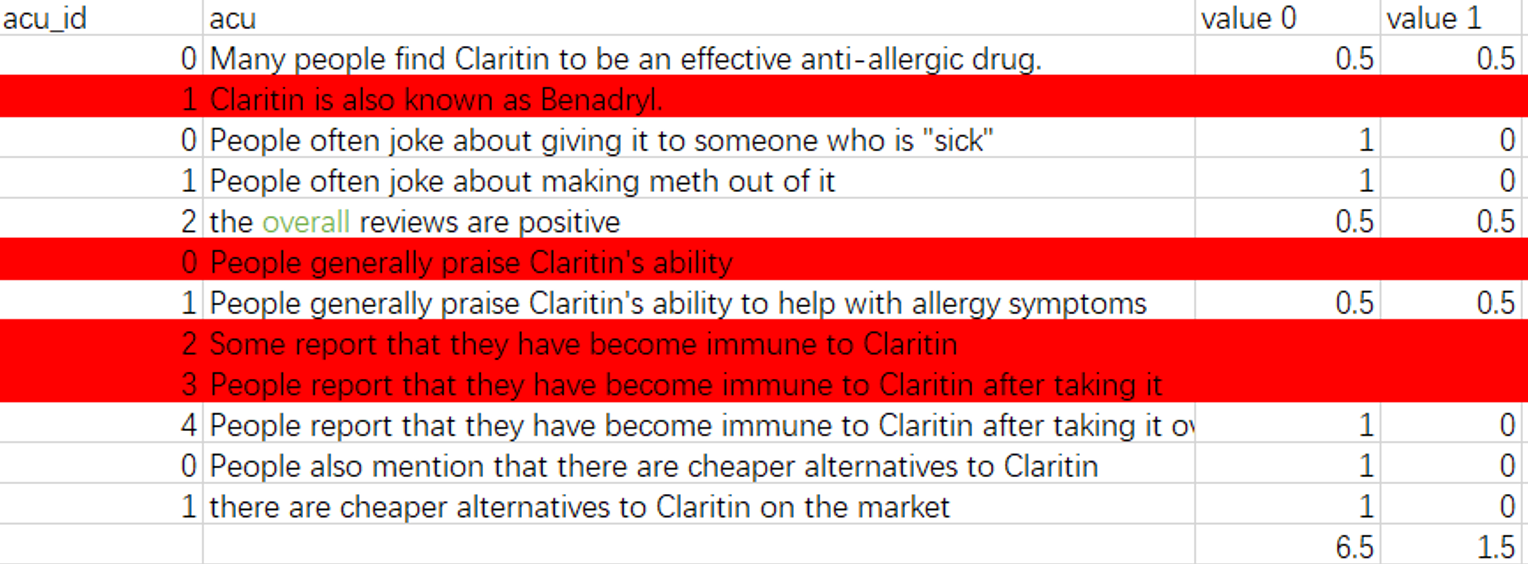}
\caption{A sample for annotation spreadsheet. The annotator can remove the ACUs (marked in red), edit the existing ACUs (marked in green), or add new ACUs during annotation. Then the annotators give scores to each value of the attribute for the ACUs. }
\label{fig:spreadsheet}
\end{figure*}

\paragraph{Sentence Fact Identification}
For this task, the annotators need to complete three steps:
\begin{itemize}
\item \textbf{Decompose summary into facts}. Please decompose the generated summary into facts. Facts in our project are countable (e.g. we can say sentence 1 contains 2 facts), usually, a longer sentence contains more facts.
 We provide a list of ACUs (Atomic Content Units) for the annotator to refer to, however, these ACUs may contain errors and inaccurate expressions. Annotators can decide how many units are there and what they are according to our own understanding of facts —-  the smallest semantic unit.
As Figure~\ref{fig:spreadsheet} shows, annotators can modify ACUs in the marked column (add, delete, or modify) to accurately show the semantic units of the sentences.
\item \textbf{Identify the source of the facts}. After getting all the facts in the summary, the next step is to identify where each fact comes from. “Comes from” here means the input source text that can fully infer (entail) this fact. We need to carefully read the input of different values, and then annotate which value the fact is generated from. There are two main cases, 1) the fact can be inferred from a single value, and 2) it is not possible to be inferred from any values (hallucination). For 1), add a counter for that value(s) (note that probably more than one value can infer this fact). For 2), add a counter for hallucination.
\item \textbf{Count the facts for all values}. After annotating the source of each fact, the program will automatically generate the count for each value. Although this step is done automatically, the annotator needs to check it to ensure fact annotation and the calculation are correct. 

\end{itemize}
\paragraph{Summary Fairness Identification}
Summary Fairness Identification only contains one step:
\begin{itemize}
\item \textbf{Rate the level of fairness}. You can refer to the ratio of facts in the spreadsheet. Give a score to rate if each value is underrepresented. We have 5 ratings for being underrepresented, if you find the ratio of one value in summary is larger or equal to 80\% of that value in source, the rate is 5. Overall, rate it according to the proportion: 5: >80\%, 4: 60\%-80\%, 3: 40\%-60\%, 2: 20\%-40\%, 1: 0\%-20\%
\end{itemize}
\paragraph{Annotation Interface}
Figure~\ref{fig:summvis} shows an example of the SummVis interface. Annotators can click on the words with underlines to see the source of the tokens to improve the annotation speed and accuracy. Finally, the results of annotations will be collected in a spreadsheet as shown in Figure~\ref{fig:spreadsheet} for further computation.

\section{Model Details}
\label{sec:model}

We compare the fairness performance of 5 models that are representative LLMs with the state-of-the-art performance on summarization tasks. 
For all models, we use a zero-shot prompt setting because LLMs have a strong capability for summarization~\citep{goyal2022zeroshotnews, zhang2023benchmarking} and we do not want the reference summaries to influence the judgment of the LLMs themselves.
The LLMs studied are listed as follows:

\paragraph{GPT} We use three models by OpenAI: two variants of GPT-3.5 (\textbf{text-davinci-003} and \textbf{gpt-3.5-turbo}) \citep{Brown2020gpt3, Ouyang2022instructgpt} and \textbf{gpt-4} \citep{openai2023gpt4}. The details regarding the training process of these models have not been published, but the authors mention that GPT-4 is trained on ``publicly available data (such as internet data) and data licensed from third-party providers''. Therefore, it is possible that these models are trained on summarization datasets.  For parameters, we set the temperature to 0 and the maximum length to 512. 

\paragraph{LLaMA 2}~\citep{touvron2023llama} is an open-sourced pretrained and fine-tuned
LLM ranging in scale from 7 billion to 70 billion parameters. We use \textbf{llama-2-7B-chat}, \textbf{llama-2-13B-chat}, and \textbf{llama2-70B-chat} which are optimized for instruction following and dialogue use cases. We set the temperature to 0 and the maximum length to 512.

\paragraph{Alpaca} \citep{alpaca} is a model fine-tuned from the LLaMA model \citep{touvron2023llama1} on 52K instruction-following demonstrations\footnote{\url{https://github.com/tatsu-lab/stanford_alpaca}}. The training data for LLaMA is acquired from seven sources, and it does not include summarization datasets. However, the training data for Alpaca includes examples of text summarization in various domains, including news articles and scientific papers. For parameters, we use a temperature of 0.1 and a context length (the maximum number of tokens from both input and output) of 2048.

\paragraph{PaLM 2}~\citep{anil2023palm} is a close-sourced LLM optimized for multilingual and reasoning tasks. It is also a Transformer-based model trained using a mixture of objectives. We directly use \textbf{text-bison@001} model provided by Google Vertex API\footnote{\url{https://cloud.google.com/vertex-ai/docs/generative-ai/model-reference/text?hl=zh-cn}} to run all experiments. We set the temperature to 0 and the maximum length to 512.

\paragraph{Claude} is an Anthropic’s model.\footnote{\url{https://docs.anthropic.com/claude/reference/selecting-a-model}} 
We use \textbf{claude-instant-1}, and the details regarding the training process of this model have not been published. Claude can help with various tasks, such as summarization, search, creative and collaborative writing, Q\&A, and coding. claude-instant-1 is one of the claude models which can consume 100k tokens.\footnote{\url{https://www.anthropic.com/index/introducing-claude}} All our experiments use their default hyper-parameters.

\begin{table*}[!ht]
\centering
\resizebox{\textwidth}{!}{

\begin{tabular}{@{}lrrrrrrrrrrrr@{}}
\toprule
& \multicolumn{2}{c}{Claritin}& \multicolumn{2}{c}{US Election}& \multicolumn{2}{c}{Yelp}& \multicolumn{2}{c}{Amazon}  & \multicolumn{2}{c}{SupremeCourt}  & \multicolumn{2}{c}{IQ2} \\ \cmidrule(lr){2-3} \cmidrule(lr){4-5} \cmidrule(lr){6-7} \cmidrule(lr){8-9} \cmidrule(lr){10-11} \cmidrule(lr){12-13}
& \multicolumn{1}{c}{BUR$\downarrow$} & \multicolumn{1}{c}{UER$\downarrow$} & \multicolumn{1}{c}{BUR$\downarrow$} & \multicolumn{1}{c}{UER$\downarrow$} & \multicolumn{1}{c}{BUR$\downarrow$} & \multicolumn{1}{c}{UER$\downarrow$} & \multicolumn{1}{c}{BUR$\downarrow$} & \multicolumn{1}{c}{UER$\downarrow$} & \multicolumn{1}{c}{BUR$\downarrow$} & \multicolumn{1}{c}{UER$\downarrow$} & \multicolumn{1}{c}{BUR$\downarrow$} & \multicolumn{1}{c}{UER$\downarrow$} \\ \midrule

\textit{\textbf{Alpaca$^\bigstar$ (7B)}}  &&&&&&&&&&&\\
N-gramScore& 51.12 & 7.63  & 82.56 & 7.01  & 34.47 & 3.71  & 74.27  &  5.11 &  70.77 &5.41& 82.14 & 4.37  \\
BERTScore& 70.85 & 12.17 & 83.27 & 6.83  & 38.73 & 5.52  & 76.20  & 6.89 &  78.46 &  6.07 & 85.94 & 6.94  \\
BARTScore& 73.84 & 11.13 & 94.31 & 9.72  & 58.93 & 6.11  &  91.80 & 9.48  & 89.23  &  7.09 & 90.78 & 10.32 \\
Average  & 65.27 & 10.31 &  86.71 & 7.85  & 44.04 & 5.11  &  80.76 &  7.16 &  79.49 & 6.19  & 86.29 & 7.21  \\ 

\textit{\textbf{llama-2-chat$^\bigstar$(7B)}}   &&&&&&&&&&&  \\
N-gramScore  & 47.56& 7.11  & 57.48 & 3.64  & 16.67 & 2.71  & 61.87 & 3.94  & 95.35 & 4.03  & 73.05 & 3.43  \\
BERTScore& 67.63& 10.70 & 84.37 & 6.28  & 47.60 & 3.09  & 73.47 & 3.95  & 96.84 & 4.42  & 87.26 & 5.72  \\
BARTScore& 73.78& 9.46  & 94.07 & 10.14 & 61.00 & 4.97  & 92.67 & 7.39  & 100.00& 6.31  & 92.93 & 8.92  \\
Average  & 62.99& 9.09  & \textbf{78.64} & 6.69  & 41.76 & 3.59  & 76.00 & 5.09  & 97.40 & 4.92  & 84.41 & 6.02  \\

\textit{\textbf{llama-2-chat$^\bigstar$(13B)}}   &&&&&&&&&&&  \\
N-gramScore  &47.04 	&7.01 	&60.07 	&3.89 	&21.80 	&3.02 	&64.33 	&4.23 	&95.34 	&4.13 	&76.21 	&3.68 \\
BERTScore&68.37 	&10.98 	&82.59 	&6.23 	&29.40 	&2.81 	&64.60 	&3.95 	&91.88 	&3.91 	&89.30 	&6.45 \\
BARTScore&73.78 	&10.07 	&95.19 	&10.26 	&57.40 	&4.76 	&91.60 	&7.16 	&99.25 	&5.70 	&92.61 	&9.85  \\
Average  &63.06 	&9.35 	&79.28 	&6.79 	&36.20 	&3.53 	&73.51 	&5.11 	&95.49 	&4.58 	&86.04 	&6.66   \\

\textit{\textbf{llama-2-chat$^\bigstar$(70B)}}   &&&&&&&&&&&   \\
N-gramScore   & 42.74& 6.60  & 66.00 & 4.16  & 22.33 & 3.26  & 64.60 & 4.43  & 95.19 & 3.96  & 75.23 & 3.59  \\
BERTScore & 66.00& 10.70 & 81.63 & 6.07  & 30.00 & 0.03  & 67.87 & 4.63  & 95.79 & 4.21  & 87.68 & 6.26  \\
BARTScore& 75.85& 12.47 & 91.70 & 8.99  & 58.80 & 5.13  & 91.60 & 7.82  & 99.85 & 6.05  & 94.23 & 10.07 \\
Average  & \textbf{61.53} & 9.92  & \textit{79.78} & 6.41  & 37.04 & 2.81  & 74.69 & 5.63  & 96.94 & 4.74  & 85.71 & 6.64  \\\bottomrule

\textit{\textbf{text-bison@001(N/A)}}   &&&&&&&&&&&   \\
N-gramScore   & 52.37& 7.77  & 74.57 & 6.11  & 32.76 & 3.66  & 75.53 & 5.52  & 96.53 & 4.29  & 81.84 & 4.50  \\
BERTScore& 72.15& 12.47 & 87.78 & 8.24  & 43.07 & 5.81  & 84.73 & 8.56  & 96.84 & 5.50  & 90.64 & 8.15  \\
BARTScore& 78.67& 11.24 & 97.23 & 11.77 & 61.11 & 6.65  & 93.00 & 11.22 & 99.85 & 7.17  & 95.29 & 11.98 \\
Average  & 67.73& 10.49 & 86.53 & 8.71  & 45.65 & 5.37  & 84.42 & 8.43  & 97.74 & 5.65  & 89.26 & 8.21  \\

\textit{\textbf{text-davinci-003 (175B)}}&&&&&&&&&&&&\\
N-gramScore& 45.41 & 6.88  & 73.04 & 5.22  & 26.87 & 2.93  & 67.80 & 4.14  & 96.09 & 4.28  & 78.11 & 3.89  \\
BERTScore& 69.48 & 11.18 & 82.30 & 6.64  & 40.27 & 4.03  & 78.27 & 5.79  & 96.09 & 4.63  & 88.67 & 6.92  \\
BARTScore& 73.93 & 9.17  & 92.89 & 9.41  & 62.13 & 6.09  & 93.60 & 9.86  & 100.00& 6.84  & 94.72 & 10.78 \\
Average & \underline{62.94} & \textbf{9.08}  & 82.74 & 7.09  & 43.09 & 4.35  & 79.89 & 6.60  & 97.39 & 5.25  & \underline{87.17} & 7.20  \\

\textit{\textbf{gpt-turbo-3.5 (175B)}} & \textit{\textbf{}}  &&&&&&&&&&&\\
N-gramScore& 51.11 & 7.53  & 72.67 & 5.00  & 21.80 & 3.01  & 63.13 & 4.12  & 96.84 & 4.16  & 72.48 & 3.37  \\
BERTScore& 68.22 & 11.04 & 80.37 & 6.10  & 33.80 & 3.70  & 71.20 & 4.93  & 92.93 & 3.76  & 86.77 & 5.68  \\
BARTScore& 73.56 & 8.98  & 91.11 & 8.35  & 60.33 & 5.30  & 93.33 & 8.40  & 100.00& 6.00  & 94.30 & 9.15  \\
Average  & 64.30 & \underline{9.18}  & \underline{81.38} & \textbf{6.48}  & \textbf{38.64} & \underline{4.00}  & \underline{75.89} & \underline{5.82}  & \textbf{96.59} & \underline{4.64}  & \textbf{84.52} & \textbf{6.07}  \\

\textit{\textbf{gpt-4 (N/A)}}&&&&&&&&&&&&\\
N-gramScore& 50.81 & 7.51  &  65.78& 4.38  & 20.27 & 2.50  & 59.87 & 3.88  & 95.34 & 4.12  & 79.45 & 3.87  \\
BERTScore& 70.44 & 11.96 & 78.81  &6.22& 36.60 & 3.21  & 72.00  &  4.39 &95.04& 3.71  & 88.88 & 6.96  \\
BARTScore& 77.85 & 10.34 &  95.19 &  10.36 & 61.40 & 5.46  &  92.47 &  8.20 &  100.00 & 5.87  & 94.79 & 10.00 \\
Average  & 66.37 & 9.94  & \textbf{79.93}  &  \underline{6.99} & \underline{39.42}& \textbf{3.72} & \textbf{74.78}  & \textbf{5.49} & \underline{96.79}  &  \textbf{4.57}& 87.71 & \underline{6.94}  \\
 \bottomrule
\end{tabular}}
\caption{Main results with our proposed metrics. 
BUR and UER are better with a lower score $\downarrow$. \textbf{Bold} indicates the best average score observed across  of all models (consistent with the metrics used in Table~\ref{tab:overall}),} and \underline{underline} indicates the second best. $^\bigstar$: Open source models.

\label{tab:overall_detail}
\end{table*}

\section{Result Details}
\label{sec:detail_scores}
Table~\ref{tab:overall_detail} shows the details of each score, including N-gramScore, BERTScore, BARTScore, and their average scores.

\section{Prompt Templates}
\label{sec:template}
For each dataset in \fairdata, we design a prompt for LLMs. For controlling sentences and adding fair prompts, we design two prompts as well. The prompt templates are listed in Table~\ref{tab:template}.

\begin{table*}[!t]
\centering
\small
\begin{tabular}{lp{13cm}}
\Xhline{5\arrayrulewidth}
  \multicolumn{2}{c}{Template for All Experiments on \fairdata}\\ \midrule
  Claritin  &   Reviews about Claritin. Each review is separated by || : \{SOURCE\} Please write a short text containing the salient information, i.e. a summary. The summary of the reviews is: \\\midrule
  US Election & Reviews about US Presidential Election. Each review is separated by || : \{SOURCE\} Please write a short text containing the salient information, i.e. a summary. The summary of the reviews is: \\\midrule
  Amazon & Reviews about a product. Each review is separated by || : \{SOURCE\} Please write a short text containing the salient information, i.e. a summary. The summary of the reviews is:\\\midrule
  Yelp & Reviews about a business. Each review is separated by || : \{SOURCE\} Please write a short text containing the salient information, i.e. a summary. The summary of the reviews is: \\\midrule
  SupremeCourt & Dialogue of the Supreme Court oral arguments. Each turn of the dialogue is one line: \{SOURCE\} The summary of the dialogue is: \\\midrule
  IQ2 & Debates on certain topics. Each turn of the dialogue is one line: \{SOURCE\} The summary of the dialogue is:\\\midrule
  \multicolumn{2}{c}{Template for Analysis on \fairdata}\\\midrule
  Sentence Control & Summary it in \{NUMBER\} sentences.\\\midrule
  Fair Instruction & \{MALE\_RATIO\}\% of the reviews are written by males and \{1-MALE\_RATIO\}\% written by females. They are mixed randomly in the source text. Please ensure the length of the male review in the summary is still \{MALE\_RATIO\}\% of the total length. \\
  
\Xhline{5\arrayrulewidth}
\end{tabular}
\caption{Prompt template on \fairdata. The first six templates are used for every sample in six datasets. The last two templates are used as additional prompts appended to the input for the use of analysis. }
\label{tab:template}
\end{table*}
\begin{table}[t!]
\centering
\begin{tabular}{@{}llll@{}}
\toprule
 & BUR$\downarrow$ & UER$\downarrow$& SOF$\downarrow$\\ \midrule
Claritin & 47.04	&7.01	&0.09 \\
\;\; + Definition &\textbf{45.26}	&\textbf{6.92}	&\textbf{0.07} \\ \midrule
Yelp &\textbf{21.80}	&\textbf{3.02}	&\textbf{1.53} \\
\;\; + Definition &23.13	&3.25	&1.78 \\ \midrule
Election &\textbf{60.07}	&\textbf{3.89}	&0.94 \\
\;\; + Definition  &68.44	&4.36	&\textbf{0.28}\\ \bottomrule
\end{tabular}
\caption{Effect of fairness instruction prompt. }
\label{tab:instruct_llama}
\end{table}
\begin{table}[]
\resizebox{\linewidth}{!}{
\begin{tabular}{@{}lccc@{}}
\toprule
model type & factuality$\downarrow$ & read. \& flu.$\uparrow$ & coherence$\uparrow$ \\ \midrule
sent=1 & 0.70   & \textbf{4.65}   & \textbf{4.75}  \\
sent=3 & 0.55   & 4.60   & 4.60  \\
natural& \textbf{0.40}   & 4.40   & 4.30  \\
sent=5 & 0.60   & 2.90   & 3.70  \\
\hline
temp=0 & \textbf{0.40}   & \textbf{4.40 }  &\textbf{ 4.30  }\\
temp=0.3   & 0.45   & 4.35   & 4.20  \\
temp=0.7   & 0.55   & 3.75   & 3.80  \\
temp=1 & \textbf{0.40}   & 4.10   & 3.70  \\
\hline
w/o definition & 0.40   & 4.40   & \textbf{4.30}  \\
w/ definition  & \textbf{0.40}   & \textbf{4.45 }  & 3.80  \\ \bottomrule
\end{tabular}}
\caption{Human evaluation on factuality, readability \& fluency, and coherence. Factuality scales from zero to one (zero is the best). The other two metrics scale from one to five, indicating the rating of the performance (five is the best).}
\label{tab:human_addition}
\end{table}
\section{Model Comparison}
\label{sec:model_comparison}
\begin{figure}[t!]
\centering
\includegraphics[width=0.95\linewidth]{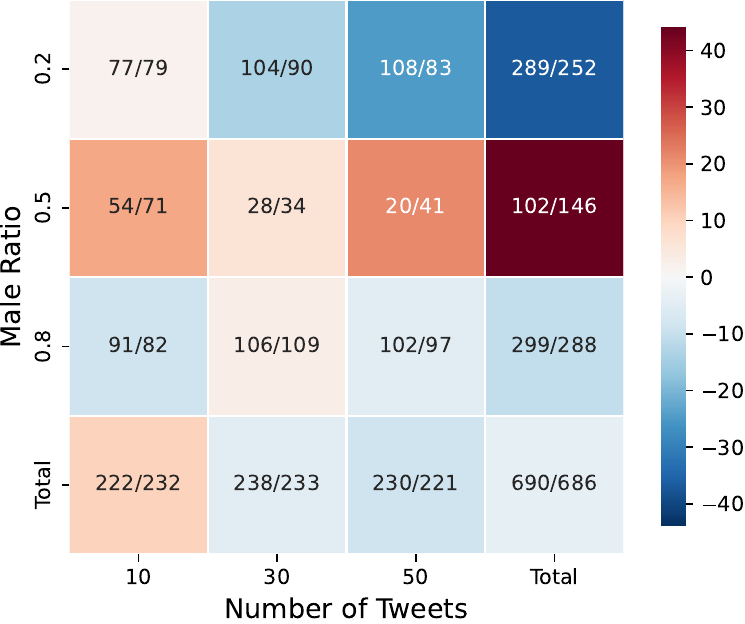}
\caption{Error comparison between gpt-turbo-3.5 and gpt-4 on Claritin. Each row shows a different male ratio while each column is a different number of tweets per sample. Each number in the cell is the number of the unfair samples generated by gpt-turbo-3.5/gpt-4. The colors of the cells become redder/bluer when gpt-turbo-3.5 produces more/less fair samples than gpt-4.}
\label{fig:model_compare}
\end{figure}

The Claritin dataset consists of different male ratios and number of tweets. We compare the performance of gpt-turbo-3.5 and gpt-4 in detail by decomposing the results into each (male ratio, number of tweets) pair. Figure~\ref{fig:model_compare} compares the gpt-turbo-3.5 and gpt-4 models in terms of the number of fairness samples on the Claritin dataset. The value $t$ in a cell means that the gpt-turbo-3.5 has $t$ more fair samples than gpt-4. As shown in the table, gpt-turbo-3.5 outperforms gpt-4 when the male ratio is 0.5 and the number of tweets is 10 while gpt-4 is better in the other cases. This shows that gpt-4 is better at processing longer tweets and the samples where the male ratio is not equal.

\paragraph{Dataset Constitution}
Figure~\ref{fig:model_compare} shows the number of error samples for each combination of male ratio and number of tweets. As can be seen, the fairness of the summaries decreases when the male ratio is not 0.5 where the ratio of male and female are the same. Besides, there is no significant difference between different numbers of tweets. 

\section{More Example Outputs}
\label{sec:example}
Table~\ref{tab:reason_case} shows two cases of model-generated summaries. Both cases turn the unfairly generated summaries into fair ones by adding fair instructions (Case 1) and by changing temperature (Case 2).
\begin{table*}[!t]
\centering
\small
\begin{tabular}{lp{13cm}}
\Xhline{5\arrayrulewidth}
\multicolumn{2}{c}{Case 1: Influence of Instructions}\\
\midrule
  Source$^\bigstar$  &  \textbf{Male Tweets}: 'mucinex and claritin. (@ cvs pharmacy) http://t.co/bju2jidt || if a girl sends you "heyy, im sick" she probably wants that claritin-d', \textbf{Female Tweets}: '2 claritin 5 aspirins and an entire bag of cough drops later and i still feel like shit || "\'if says she\'s sick, she probably wants the d.\' claritin d?" - mom  hahahaha i love my mom!!! || my mom said i was grounded because i\'m threatening to overdose on claritin. || i woke wit these allergies bout to pop sme claritin and get to wrk || i took a claritin and motrin 30 minutes ago! || i have claritin rt @juandelay: anybody got some oatmeal or benadryl?? im havin a allergic || @jonathanrknight i prefer claritin, works better for me. || there\'s clear, and then there\'s claritin clear'\\
  \midrule

  w/o instruction & The reviews about Claritin, an anti-allergic drug, are mixed. Some people find it effective in relieving their allergies, while others mention using it as a joke or experiencing negative side effects. Overall, the reviews highlight different experiences and opinions regarding the use of Claritin.\\ \midrule
  w/ instruction & Claritin is a popular anti-allergic drug that seems to work well for many people. Some individuals use it to alleviate allergies and get through their day, while others joke about its potential effects. Overall, the reviews indicate that Claritin is effective in providing relief from allergy symptoms. \\

   \Xhline{5\arrayrulewidth}
\multicolumn{2}{c}{Case 2: Influence of Temperature}\\
\midrule
  Source$^\bigstar$ & \textbf{Male Tweets}: if a girl says she's sick. give her the claritin d. || @bass\_courtney: @21ccharlesworth oh my! yess i need some claritin -d.:p @shewantyourdick  add her to the d list \textbf{Female Tweets}: @sawngbyrd28 yes you can overdose on claritin.  any signs of muscle contractions, drowsiness, or headache? || @kamandermainy: this sinus pressure is not the business i just got over mine.... don't u hate that? claritin d will clear it right up tho || just took this claritin... no snoring for me tonight lol. || in big trouble. forgot to buy claritin-d. may have to get a hotel for the night. \#madhusbandwhenigethomeemptyhanded || @4jinxremoving i'm already on claritin and nasonex!!! || @so\_freshgiraffe : actually, i'm allergic to cats. you know; red eyes, sneezing, hives. but i can't stay away from him. claritin helps (: || smh it never fails every time i come home my allergies start actin up!!---but its all good cause im prepared this time! *pulls out claritin* || @morethymeblog hi! good day today. took 1/2 of a claritin bcz allergies has been acting up on top of everything. my mom's b'day today! you? \\ \midrule
  Temp=0 & People have found relief from allergies and sinus pressure using Claritin, with some even avoiding snoring at night. However, there are concerns about overdosing on the medication, which can cause muscle contractions, drowsiness, or headaches. \\ \midrule
  Temp=1 & Some Twitter users mentioned needing Claritin to relieve allergies, particularly sneezing, red eyes, and sinus pressure. The drug appears to be effective in providing relief for these symptoms. It is mentioned that overdosing on Claritin is possible, which can lead to muscular contractions, drowsiness, and headache. Overall, many of the reviewers expressed satisfaction with Claritin's effectiveness in managing their allergies.\\
   \Xhline{5\arrayrulewidth}

\end{tabular}
\caption{Two case studies on Claritin. Predictions are generated by gpt-turbo-3.5. After adding fair instructions, the generated summary becomes fair (BUR becomes zero). After tuning the temperature from zero to two, the generated summary becomes fair (BUR becomes zero). $\bigstar$ For the input to the model, male/female are not marked and all tweets are randomly shuffled; this table shows these gender values for the purpose of clarity.} 
\label{tab:reason_case}
\end{table*}

Table~\ref{tab:reference} shows two samples of reference summaries on Yelp and Amazon datasets. In sample one, the reference summary ignores the negative reviews which is important for the users to gain a comprehensive understanding of the product. For sample two, the summary overly emphasizes the negative aspects while disregarding the review that awarded a five or four-star rating. In both examples, our metrics successfully capture that these samples are unfair, as well as underrepresenting which values.

\begin{table*}[!t]
\centering
\small
\begin{tabular}{lp{13cm}}
\Xhline{5\arrayrulewidth}
\multicolumn{2}{c}{Sample One from Yelp Dataset}\\
\midrule
  Source$^\bigstar$  &  \textbf{Positive Reviews}: Well the atmosphere is excellent - especially in CU. The staff was all very friendly and attentive. Kudos to them! The food was a bit better than average but nothing to knock your socks off. A good place to go with family or friends.They have a cheese sauce that is a bit spicy but tasty! || Great location and atmosphere. Authentic and Mexican American fare that was served in generous portions and DELICIOUS! The Guadalajara burrito was simple and SO flavorful. Hubs had Molcalambres and ranked them near the top of his fajita list. Not hoiity toity; not trying to be hip and trendy like Maize. Will come back whenever in town. || Love this place! The waitstaff is great and friendly, they are very quick but also allow you to relax and not feel rushed. I go here in the summer with some friends and we just sit on the patio (great outdoor seating) and drink margaritas and enjoy the complimentary fresh chips and salsa. Also great food! || They took a while to bring the food, my son ordered a chicken quesadilla that was very soggy and tasteless, the rest of the food was edible but far from good, place was ok, it's far from the best but if you hungry and desperate you can certainly go eat there. || Good food at a great price, the restaurant has become a great lunch destination. The food is basic American-Mexican combinations (burrito, enchilada, tamale for example) with very quick service. Honestly have not been to a sit down restaurant with such speedy service during the lunch rush. || Their food is decent, but the reason I'm writing this review is their healthy margarita. It's incredible and I'm not even usually a margarita fan! My husband ordered the regular margarita and we both agreed that the healthy was much better than the regular margarita. Go try it! || Went here with a group to drink some margaritas and eat some chips and salsa. Wonderful service and friendly staff. I ordered a grande raspberry margarita and it was nice and strong. Chips were hot and delicious and salsa was good (wish it was spicier but that's a personal preference),
  \textbf{Negative Reviews} The first time I went here, I ordered a taco salad with beef, and the \underline{beef was pink}. Not brown on the outside and pink on the inside, pink pink pink. The second time I went, my friend and I had the same burrito combination platter and \underline{both of us got sick} the next day. \\ \midrule

  Reference Summary & This Mexican restaurant is a great place to get a margarita with chips and salsa. Their margaritas are very well done, and the chips and salsa are also good. The service is also pretty good, friendly and efficient. The rest of the food, however, is average at best. This restaurant is recommended mostly for having some drinks with friends.
 \\

   \Xhline{5\arrayrulewidth}
\multicolumn{2}{c}{Sample Two from Amazon Dataset}\\
\midrule
  Source$^\bigstar$ & \textbf{Rating=1} I'm 5 '4 'and this tank fits like a normal tank top, not any longer. I was trying to find a tank that would cover past my hips, so I could wear it with leggings. Don't order if you're expecting tunic length.
\textbf{Rating=2} The only reason I'm rating this at two stars is because it is listed as a 'long' tank top and the photo even shows it going well past the models hips, however I'm short and the tank top is just a normal length.
\textbf{Rating=3} The description say it long... NOT so it is average. That's why I purchased it because it said it was long. This is a basic tank.I washed it and it didn't warp but did shrink a little. Nothing to brag about. || This shirt is OK if you are layering for sure. It is THIN and runs SMALL. I usually wear a small and read the reviews and ordered a Medium. It fits tight and is NOT long like in the picture. Glad I only purchased one. || I usually get them someplace out but they no longer carry them. I thought I would give these a try. I received them fast, although I did order a brown and got a black (which I also needed a black anyway). They were a lot thinner than I like but they are okay.
\textbf{Rating=4} The tank \underline{fit very well and was comfortbale to wear}. The material was thinner than I expected, and I felt it was probably a little over priced. I've bought much higher quality tanks for \$5 at a local store. || I bought this tank to wear under shirts when it is colder out. I bought one in white and one in an aqua blue color. They are long enough that the color peeks out from under my tops. \underline{Looks cute}. I do wish that the neck line was a bit higher cut to provide more modest coverage of my chest.
 \textbf{Rating=5}  Every women should own one in every color. \underline{They wash well perfect under everything}. Perfect alone. As I write I'm waiting on another of the same style to arrive. Just feels \underline{quality} I don't know how else to explain it, but I'm sure you get it ladies!
 \\ \midrule
  Reference summary & Some Twitter users mentioned needing Claritin to relieve allergies, particularly sneezing, red eyes, and sinus pressure. The drug appears to be effective in providing relief for these symptoms. It is mentioned that overdosing on Claritin is possible, which can lead to muscular contractions, drowsiness, and headache. Overall, many of the reviewers expressed satisfaction with Claritin's effectiveness in managing their allergies.\\
   \Xhline{5\arrayrulewidth}

\end{tabular}
\caption{Two samples of unfair human-written reference summaries. Sample one is from Yelp dataset and ignores the negative review. Sample two is from Amazon dataset and ignores reviews with four or five stars. \underline{Underline} words indicate the aspects that the reference summary ignores. $\bigstar$ For the input to the model, sentiment/ratings are not marked and all reviews are randomly shuffled; this table shows these values for clarity.} 
\label{tab:reference}
\end{table*}
\section{Analysis on Fairness Improvement}
We conduct a human evaluation task assessing the factuality, readability, fluency, and coherence of the proposed approaches, incorporating diverse lengths, temperatures, and fair prompts.

Specifically, we ask annotators to annotate 20 summaries from Claritin dataset generated by gpt-turbo-3.5 with various settings. For each summary, the annotator needs to annotate three metrics: For factuality, annotate one if any factuality errors are found, otherwise zero. For readability \& fluency, and coherence, the annotators need to rate from one to five to indicate the level of performance (five indicates the best performance). Table~\ref{tab:human_addition} shows the average score of each summary in various settings.

As shown in the table, readability, fluency, and coherence decrease with temperature and length increasing. For factuality, experiments show that the proposed approaches do not significantly hurt factuality.

\section{Fairness Instruction on LLaMA}
We conducted an experiment with the llama2-13b model, revealing that after adding the definition, fairness increased for Claritin dataset but decreased for Yelp and Election (Table~\ref{tab:instruct_llama}). We speculate that the impact of adding a fairness prompt is linked to the instruction-following capability of LLMs. To validate our hypothesis, we conducted further qualitative analysis, finding that the results of LLaMA 2 contained noises after adding a fairness definition, such as repeating the question. For instance, redundant sentences like “The generated summary will be more fair…” are found in generated summaries.

\end{document}